\documentclass[twoside,11pt]{article}
\usepackage[T1]{fontenc}
\usepackage{amsmath,amssymb,amsthm,mathrsfs,accents}
\usepackage{thmtools,thm-restate}
\let\originalproof\proof
\let\endoriginalproof\endproof
\usepackage[preprint]{jmlr2e}
\let\proof\originalproof
\let\endproof\endoriginalproof
\usepackage{algorithm,algorithmic,enumitem,longtable,booktabs,array,calc}
\usepackage{pgfplots}
\pgfplotsset{compat=1.18}

\newcommand{\ep}{\epsilon}

\newcommand{\fisher}{\mathcal{I}}

\newcommand{\ex}{\mathbb{E}}

\newcommand{\dlim}{\xrightarrow{d}}
\newcommand{\defn}{\triangleq}

\ShortHeadings{Private Confidence Regions for M-Estimation}{Chen, Cho and Awan}
\begin{document}
\title{Private Confidence Regions for M-Estimation via Exponential Mechanism Asymptotics}
\author{\name Yu-Wei Chen \email chen4357@purdue.edu \\
        \addr Department of Statistics, Purdue University\\
        West Lafayette, IN, USA
        \AND
        \name Young Hyun Cho \email younghyuncho@fas.harvard.edu \\
        \addr Department of Statistics, Harvard University \\
        Cambridge, MA, USA 
        \AND
        \name Jordan Awan \email jaa557@pitt.edu \\
        \addr Department of Statistics, University of Pittsburgh\\
        Pittsburgh, PA, USA}
\editor{}
\maketitle

\begin{abstract}
The exponential mechanism is a fundamental differential privacy mechanism, which can be used for private M-estimation. Recent work has extended it to satisfy Gaussian differential privacy (GDP) with improved dependence on dimension and sample size, motivating a deeper understanding of its statistical behavior. We develop a general asymptotic framework for M-estimation under the GDP exponential mechanism, revealing how statistical uncertainty interacts with the randomness introduced by the exponential mechanism. Our analysis shows that privacy-induced randomness is asymptotically negligible and matches minimax lower bounds under strongly convex loss, while characterizing the intrinsic trade-off between privacy and regularization in the general convex setting. We further establish that sampling variability and mechanism-induced randomness are asymptotically independent, yielding a general limit theorem for randomized post-data procedures. We apply these results to construct asymptotically valid confidence regions with a finite-sample correction. Simulation studies and real-data experiments demonstrate that the proposed method achieves nominal coverage more rapidly than competing differentially private inference methods, highlighting both its theoretical guarantees and practical effectiveness.

\end{abstract}


\section{Introduction}\label{sec: intro}
Sensitive data has been increasingly collected and analyzed to produce accurate estimates and to inform decisions, which has raised concerns over privacy leaks. Differential privacy (DP; \citealp{dwork2006calibrating}) provides a rigorous framework for limiting what a released analysis reveals about any individual observation. Its adoption in industry and the public sector \citep{erlingsson2014rappor,team2017learning,ding2017collecting,abowd20222020} has increased the need for statistical inference methods that account for both the data randomness and the noise introduced for privacy protection. 

The exponential mechanism (ExpMech; \citealp{mcsherry2007mechanism}) is a general method for differentially private selection. 
It samples from the output space with a distribution determined by a utility score, favoring outputs with higher utility. When applied to empirical risk minimization (ERM; \citealp{chaudhuri2011differentially,kifer2012private,bassily2014private}), the ExpMech uses the negative empirical loss as its utility score, assigning greater density to parameters with smaller loss. Its privacy calibration controls how tightly this distribution concentrates around the empirical minimizer and thus affects the statistical behavior of the output.

Although the initial calibration of the ExpMech guarantees pure DP, its mechanism error can be comparable to sampling error and grow linearly in dimensions \citep{awan2019benefits}. Recent work by \citet{gopi2022private} uses strong convexity of the loss to calibrate the ExpMech for Gaussian differential privacy (GDP; \citealp{dong2022gaussian}), obtaining improved excess-risk guarantees in terms of both dimension and sample size. Their calibration incorporates the curvature of the loss as well as a Lipschitz bound on loss differences across observations. For losses that are convex but not strongly convex, \citet{gopi2022private} introduce a regularizer to ensure strong convexity, which introduces a bias.



The GDP calibration of exponential mechanism is promising, but the results of \citet{gopi2022private} are limited to excess-risk analysis, rather than distributional results. Thus, the goal of this paper is to 
study parameter inference based on GDP-calibrated ExpMech estimators. We develop an asymptotic theory for these private M-estimators under strongly convex and regularized convex losses. We characterize the mechanism error while allowing the privacy budget to vary with sample size and derive a correction for the leading regularization bias. We also establish a general joint convergence result for the sampling and mechanism errors, enabling us to construct confidence regions for the parameters that account for both sources of uncertainty. Simulations and a real-data analysis demonstrate improved finite-sample coverage compared to alternative private inference methods.

\subsection{Our Contributions}

Our analysis makes three contributions to the statistical inference of the ExpMech.

\textbf{Asymptotic Results for the ExpMech.}
{ We develop central limit theorems (CLTs) characterizing the mechanism error of the ExpMech under both strongly convex and convex losses, under reguarity conditions. In the strongly convex case, the mechanism error matches minimax lower bounds for benchmark problems \citep{cai2021cost}. When the loss is merely convex, the scale of the regularizer affects the asymptotic behavior and gives three asymptotic regimes: (i) variance-dominated, (ii) balanced, and (iii) bias-dominated. While the ExpMech-induced noise is asymptotically negligible compared to the sampling error, it is not able to attain benchmark minimax lower bounds due to the regularization. }

\textbf{Regularization Bias and Correction.}
{For convex losses, regularization is needed to guarantee strong convexity, tightening the concentration of the ExpMech; however, it also introduces bias. In the case that the bias and variance are balanced, the bias term is not asymptotically negligible and must be corrected. We develop a first-order bias correction in this case, which we show enables consistent estimation. }

\textbf{Joint Private Inference.}
{While the above asymptotic results study the randomness in the ExpMech, valid statistical inference must account for both the data randomness and the mechanism randomness. We prove that the asymptotic distribution of the ExpMech is ``asymptotically independent'' of the standard limiting distribution of the non-private M-estimator, allowing us to combine these distributions to produce asymptotic confidence regions. 
Our analysis may apply to other procedures that introduce randomness after observing the data, such as bootstrap methods or other privacy mechanisms. }

\subsection{Related Literature}

Our work relates to three lines of research: the ExpMech, uncertainty quantification under DP, and, more broadly, private optimization.

\textbf{Exponential Mechanism.}
The ExpMech has been used for various tasks, including ERM \citep{mcsherry2007mechanism,chaudhuri2011differentially,bassily2014private}. Other applications include quantile estimation \citep{smith2011privacy,gillenwater2021differentially}. Earlier studies of exponential and related mechanisms establish limiting distributions conditional on fixed data sequences \citep{awan2019benefits,reimherr2019kng}. The strong-convexity calibration of \citet{gopi2022private} provides GDP guarantees and excess-risk bounds. Our analysis develops parameter-error limits under this calibration and combines them with sampling variability for inference.

\textbf{Uncertainty Quantification under DP.}
One line of research develops statistical inference by analyzing specific DP mechanisms \citep{wang2018differentially,avella2023differentially,wang2025differentially,xia2025inference}. A complementary literature develops inference through private hypothesis testing \citep{awan2018differentially,awan2020binomial,awan2023canonical}, simulation-based methods \citep{awan2025simulation,wang2025optimal}, and private posterior samplers \citep{gong2022exact,ju2022data,chen2025particle}. We develop mechanism-specific asymptotics for the GDP ExpMech alongside a general result for combining conditional randomization limits with sampling limits. Additional background appears in Appendix~\ref{appendix: more background}.

\textbf{Private Optimization.}
Objective and output perturbation, introduced for private ERM by \citet{chaudhuri2011differentially} and refined by \citet{kifer2012private}, add noise to the optimization objective or the optimized estimate. Noisy gradient methods perturb the optimization trajectory \citep{bassily2014private,abadi2016deep}, with subsequent work analyzing their inferential properties as discussed above. Robust statistics and influence-function expansions provide another route to private estimation with controlled sensitivity \citep{avella2021privacy}. The ExpMech instead specifies the released estimator through a distribution determined by the empirical objective, making its concentration and regularization central to the analysis.

\subsection{Organization}
The remainder of the paper is organized as follows. Section~\ref{sec: background} introduces the background and problem setup. Section~\ref{sec: conditional asymptotics} develops conditional asymptotics for strongly convex and regularized convex losses. Section~\ref{sec: joint asymp} establishes joint asymptotics and constructs private confidence regions. Section~\ref{sec: simulation} presents simulations and a real-data analysis, and Section~\ref{sec: discussion} concludes. Proofs, supplementary results, and implementation details appear in the Appendix. The source code for our simulations and empirical analysis is available at \url{https://github.com/y-w-chen/AsympUQ_RegExpMech}.

\section{Background and Problem Setup}\label{sec: background}

We introduce M-estimation and the hypothesis-testing formulation of DP, then describe the ExpMech calibration and the inferential problem studied in this paper.

\subsection{M-Estimation}\label{sec: m estimation setup}

Let $D_n=(s_1,\ldots,s_n)$ be an i.i.d.\ sample from a distribution $\Pi$ on $\mathcal S$, and let $f_\theta:\mathcal S\to\mathbb R$ be a loss indexed by $\theta\in\Theta\subseteq\mathbb R^d$. The population parameter of interest is defined as,
\[
\theta_0=\arg\min_{\theta\in\Theta}F(\theta),
\]
where $ F(\theta)=\mathbb E_{S\sim\Pi}[f_\theta(S)]$. Its empirical counterpart is,
\begin{equation}\label{eq: empirical loss}
    \hat\theta_n\in\arg\min_{\theta\in\Theta}F_n(\theta;D_n),
\end{equation}
where $F_n(\theta;D_n)=\frac1n\sum_{i=1}^n f_\theta(s_i)$.
This formulation includes maximum likelihood estimation and more general M-estimators \citep{huber1964robust,van2000asymptotic}. We use the loss-minimization convention throughout and suppress the dependence of $F_n$ on $D_n$ for simplicity. 
Throughout, $\|\cdot\|$ denotes the Euclidean norm for vectors and the induced operator norm for matrices.

Under suitable regularity conditions, $\hat\theta_n$ in \eqref{eq: empirical loss} is consistent for $\theta_0$ and satisfies
\[
\sqrt n(\hat\theta_n-\theta_0)\xrightarrow{d}N_d(0,\Sigma_1),
\]
where $\Sigma_1=H^{-1}\mathcal I H^{-1}$, $H=\mathbb E[\nabla^2 f_{\theta_0}(S)]$ and $\mathcal I=\mathbb E[\nabla f_{\theta_0}(S)\nabla f_{\theta_0}(S)^\top]$. Here $H\succ0$ describes the curvature of the population loss, while $\mathcal I$ describes score variability. For a likelihood that satisfies the information equality, $\mathcal I=H$ and $\Sigma_1=H^{-1}$. Section~\ref{sec: joint asymp} states the population assumptions used in our analysis.

{ The empirical optimization problem in \eqref{eq: empirical loss} is the basis of ERM, where the goal is to minimize the gap between $F(\hat\theta)$ and $F(\theta_0)$, rather than to directly make inferences about $\theta_0$. Note that in ERM, regularization can be beneficial to reduce the average loss, while in M-estimation, regularization can introduce bias and complicate limiting distributions. } 

\subsection{Differential Privacy}\label{sec: privacy background}

A randomized mechanism $\mathcal M$ maps a dataset to a distribution over possible outputs. DP constrains how much this distribution can change when one observation is replaced. We say two datasets $D,D'\in\mathcal S^n$ are neighboring if they differ in at most one observation, and denote them as $D\sim D'$.

We use the hypothesis-testing formulation of DP \citep{dong2022gaussian}. For distributions $P$ and $Q$, define the tradeoff function for $\alpha\in[0,1]$ as
\[
T(P,Q)(\alpha)=\inf_{\phi}\left\{1-\mathbb E_Q[\phi]:\mathbb E_P[\phi]\le\alpha\right\},
\]
where $\phi$ ranges over measurable test functions taking values in $[0,1]$. Thus, $T(P,Q)(\alpha)$ is the smallest Type~II error attainable when testing $P$ against $Q$ with Type~I error at most $\alpha$.

\begin{definition}[$f$-DP, \citealp{dong2022gaussian}]
Let $f$ be a tradeoff function. A randomized mechanism $\mathcal M$ satisfies $f$-DP if, for every neighboring pair $D,D'$ and $\alpha \in [0,1]$,
\[
T\bigl(\mathcal M(D),\mathcal M(D')\bigr)(\alpha)\ge f(\alpha).
\]
\end{definition}

When $f=T(P,Q)$, this guarantee means that distinguishing mechanism outputs on neighboring datasets is at least as difficult as distinguishing $P$ from $Q$. The framework includes pure $\varepsilon$-DP \citep{dwork2006calibrating}, which corresponds to $f_\varepsilon(\alpha)=\max\{0,\,1-e^\varepsilon\alpha,\,e^{-\varepsilon}(1-\alpha)\}.$ Gaussian differential privacy (GDP) uses the benchmark
\[
    G_\epsilon(\alpha)
    =T\bigl(N(0,1),N(\epsilon,1)\bigr)(\alpha)
    =\Phi\bigl(\Phi^{-1}(1-\alpha)-\epsilon\bigr),
\]
where $\Phi$ is the standard normal cumulative distribution function. A mechanism is $\epsilon$-GDP if it is $G_\epsilon$-DP. A smaller $\epsilon$ gives stronger privacy protection. We use GDP as our primary privacy framework and allow a positive deterministic budget $\epsilon_n$ to vary with sample size.

Two properties will be used throughout.
\begin{itemize}
    \item \textbf{Post-processing.} Let $\mathcal M:\mathcal S^n\to\mathcal Y$ be a randomized mechanism, and let $\mathrm{Proc}:\mathcal Y\to\mathcal Z$ be a measurable, possibly randomized, map that does not depend on the data. If $\mathcal M$ is $f$-DP, then $\mathrm{Proc}\circ\mathcal M$ is $f$-DP.
    \item \textbf{Composition.} Suppose that $M_1$, $M_2$, $\ldots$, $M_k$ satisfy $\ep_1$-GDP, \ldots, $\ep_k$-GDP, respectively, where each mechanism takes in the dataset and possibly the output of the mechanisms with lesser indices. Then the joint release of all $k$ mechanisms satisfies $\lVert (\ep_1,\ldots, \ep_k)^\top \rVert$-GDP. 
\end{itemize}

{ The Gaussian mechanism can be used to directly add noise to a statistic, which is scaled to its sensitivity: }
\begin{proposition}[Gaussian mechanism, \citealp{dong2022gaussian}]\label{prop:gdp_mechanism}
Let $q:\mathcal S^n\to\mathbb R^p$ have finite $\ell_2$-sensitivity $\Delta_2q:=\sup_{D\sim D'}\|q(D)-q(D')\|_2.$
For $\epsilon>0$, the mechanism
\[
\mathcal M(D)=q(D)+Z,    
\]
where $Z\sim N_p\!\left(0,\frac{(\Delta_2q)^2}{\epsilon^2}I_p\right)$ is independent of the data, satisfies $\epsilon$-GDP.
\end{proposition}
{ We will apply the Gaussian mechanism to estimate the asymptotic covariance of our private M-estimators. }
\subsection{Private M-Estimation via the ExpMech}\label{sec: private m estimation}

The ExpMech converts an empirical loss into a distribution over the parameter space $\Theta$. Conditional on $D_n$, it releases $\theta_n^{DP} \in \Theta$ with density
\begin{equation}\label{equ: exp mech}
    \pi_n(\theta\mid D_n)
    =\frac{\exp\{-k_nF_n(\theta;D_n)\}}
    {\int_\Theta\exp\{-k_nF_n(u;D_n)\}\,du},
\end{equation}
provided that the denominator is finite and positive. Larger $k_n$ concentrates the distribution more tightly around the empirical minimizer. Privacy calibration determines how large $k_n$ can be for a prescribed guarantee.

The conventional $\varepsilon$-DP calibration uses a uniform bound on the change in objective values. If
\[
    \Delta F_n
    =\sup_{D\sim D'}\sup_{\theta\in\Theta}
    |F_n(\theta;D)-F_n(\theta;D')|<\infty,
\]
then choosing $k_n=\varepsilon/(2\Delta F_n)$ gives $\varepsilon$-DP \citep{mcsherry2007mechanism}, provided that $\Delta F_n>0$ and \eqref{equ: exp mech} is a well-defined density. On the other hand, the GDP calibration of \citet{gopi2022private} requires strong convexity and a Lipschitz bound on loss differences: 

\begin{restatable}[GDP guarantee for the ExpMech, \citealp{gopi2022private}]{proposition}{restatePropDpGuaranteeOfExpMech}\label{prop: dp guarantee of exp mech}
Suppose $\Theta$ is convex and $F_n(\cdot;D)$ is $\mu$-strongly convex on $\Theta$ for every dataset $D\in\mathcal S^n$, with a common $\mu>0$. Suppose also that, for observations $s\sim s'$, the map $\theta\mapsto f_\theta(s)-f_\theta(s')$ is $L$-Lipschitz on $\Theta$, with a common $L>0$. For any $\epsilon_n>0$, the ExpMech in \eqref{equ: exp mech} with $k_n={\epsilon_n^2\mu n^2}/{L^2}$ satisfies $\epsilon_n$-GDP, provided its density is well-defined.
\end{restatable}

When $F_n$ is convex, an $\ell_2$ regularizer ensures strong convexity. For a positive deterministic sequence $\mu_n$, let
\begin{equation}\label{eq: regularized empirical loss}
    F_n^\#(\theta;D_n)
    =F_n(\theta;D_n)+\frac{\mu_n}{2n}\|\theta\|^2,
    \qquad
    \theta_n^\#=\arg\min_{\theta\in\Theta}F_n^\#(\theta;D_n).
\end{equation}
The regularized objective $F^\#_n$ is $\mu_n/n$-strongly convex. Because the regularizer is data-independent, it cancels in the difference between neighboring objectives and does not affect the Lipschitz constant $L$. The proposition therefore gives $\epsilon_n$-GDP for the regularized exponential mechanism (RegExpMech), with density
\begin{equation}\label{equ: reg exp mech}
    \pi_n(\theta\mid D_n)\propto\exp\{-k_nF_n^\#(\theta;D_n)\},
\end{equation}
where $k_n=\frac{\epsilon_n^2\mu_n n}{L^2}$.

The calibration above was developed to control excess risk \citep{gopi2022private}. For inference on $\theta_0$, we must also account for the bias introduced by regularization and the sampling variability of $\hat\theta_n$. The total error decomposes as
\begin{equation}\label{eq: total error decomposition}
    \theta_n^{DP}-\theta_0
    =(\hat\theta_n-\theta_0)
    +(\theta_n^\#-\hat\theta_n)
    +(\theta_n^{DP}-\theta_n^\#).
\end{equation}
The three terms on the right-hand side are the sampling error, regularization bias, and mechanism error, respectively.
The middle term is the data-dependent displacement that generates regularization bias in our asymptotic expansions. When the original loss is strongly convex and no regularizer is used, this term is absent. The following sections characterize the three components and their joint behavior to obtain inference for $\theta_0$.

\begin{remark}
{ The GDP calibration in Proposition \ref{prop: dp guarantee of exp mech} is qualitatively different from the conventional ExpMech, as it uses different aspects of the loss function to calibrate the value of $k_n$. In particular, \citet{chen2026optimal} derived the tight privacy guarantee of the conventional ExpMech in terms of GDP, which converts the privacy parameters, but does not enjoy the improved dimensional and sample size scaling as demonstrated in \citet{gopi2022private}.}
\end{remark}

\section{Conditional Asymptotics} \label{sec: conditional asymptotics}

Using the setup in Section~\ref{sec: background}, we state the assumptions for the conditional analysis and establish central limit theorems for ExpMech's error under convex and strongly convex losses.

\subsection{Assumptions for Conditional Analysis}\label{sec: assumptions}

In this section, we hold the dataset sequence $(D_n)_{n\ge1}$ fixed and consider the ExpMech and RegExpMech defined in Section~\ref{sec: private m estimation}. All convergence statements in this section concern mechanism randomness alone.

    

Let $\zeta\in(0,1/10)$, which is held fixed throughout Sections~\ref{sec: conditional asymptotics} and~\ref{sec: joint asymp}. We impose the following conditions on the fixed dataset sequence.

\begin{enumerate}
    \item[\textbf{(A1)}] \textbf{Parameter Space.} The parameter space $\Theta \subset \mathbb{R}^d$ is convex, and the true parameter $\theta_0$ is in the interior of $\Theta$.


    \item[\textbf{(A2)}] \textbf{Loss-function Smoothness.}
For each observation $s_i$ in the fixed dataset sequence, $f_\theta(s_i)$ is twice continuously differentiable on an open neighborhood $\mathcal U\subset\operatorname{int}(\Theta)$ of $\theta_0$. There is a finite nonnegative measurable function $M$ such that $\|\nabla^2f_\theta(s_i)-\nabla^2f_{\theta'}(s_i)\|\le M(s_i)\|\theta-\theta'\|$ for all $\theta,\theta'\in\mathcal U$, and $\sup_n n^{-1}\sum_{i=1}^nM(s_i)<\infty$. In addition, for every pair $s,s'\in\mathcal S$, the map $\theta\mapsto f_\theta(s)-f_\theta(s')$ is $L$-Lipschitz on $\Theta$, with a finite constant $L$.

    \item[\textbf{(A3)}] \textbf{Hessian Convergence.}
There is a fixed positive-definite matrix $H$ such that $n^{1/2-\zeta}\|\nabla^2F_n(\theta_0)-H\|\longrightarrow0$.

    \item[\textbf{(A4)}] \textbf{Minimizer Rate and Consistency.}
Along the fixed dataset sequence, $n^{1/2-\zeta}\|\hat\theta_n-\theta_0\|\to0$. In the regularized case, for the prescribed positive deterministic sequence $\mu_n=o(n)$, also assume $\theta_n^\#\to\theta_0$.
\end{enumerate}

Note that in (A1), the parameter space need not be bounded: strong convexity, either assumed directly or supplied by the quadratic regularizer, makes the sampling objective grow at least quadratically, so its exponential has Gaussian tail decay and a finite normalizing constant. Assumptions (A3) and (A4) control the convergence rates of the empirical Hessian at $\theta_0$ and the empirical minimizer, respectively. Together with the smoothness condition in (A2), they allow us to quantify the regularization bias. The loss-difference Lipschitz condition in (A2) provides the constant $L$ used for privacy calibration.

In particular, by (A2), for $\theta\in\mathcal U$, we have
\begin{equation}\label{eq: local Hessian control}
\|\nabla^2F_n(\theta)-H\|\le\|\nabla^2F_n(\theta_0)-H\|+\left(\frac1n\sum_{i=1}^nM(s_i)\right)\|\theta-\theta_0\|.
\end{equation}
Together with (A3), this implies that $\nabla^2F_n(\theta_n)\to H$ whenever $\theta_n\to\theta_0$. This local Hessian control is used in the conditional asymptotic results below. 

\subsection{Conditional Asymptotics for DP Randomness in ExpMech}\label{sec: dp error asymp}

Under (A1)--(A4), we first analyze strongly convex losses and then quantify the mechanism noise and regularization bias for convex losses. This conditional approach follows the fixed-dataset analyses of \citet{awan2019benefits} and \citet{reimherr2019kng}.


\begin{restatable}[CLT for GDP ExpMech under strongly convex loss]{theorem}{restateThmStrongConvUtility} \label{thm: strong conv utility}
Suppose (A1)--(A4) hold and each $F_n$ is $\mu$-strongly convex on $\Theta$ for a common $\mu>0$. Let $\epsilon_n n\to\infty$. For the ExpMech in \eqref{equ: exp mech}, calibrated as in Proposition~\ref{prop: dp guarantee of exp mech}, we have
\[
\epsilon_n n(\theta_n^{DP}-\hat\theta_n)
\dlim N_d\!\left(0,\frac{L^2}{\mu}H^{-1}\right),
\]
where convergence is over the mechanism randomness, with the dataset sequence held fixed.
\end{restatable}

Theorem~\ref{thm: strong conv utility} gives mechanism error of order $1/(\epsilon_n n)$, reducing to $1/n$ for a fixed positive privacy budget. Its ratio to sampling error is of order $1/(\epsilon_n\sqrt n)$, so it is negligible at the statistical scale when $\epsilon_n\sqrt n\to\infty$. For comparison, under $(\varepsilon,\delta)$-DP with $\varepsilon\le1$ and sufficiently small $\delta$, the minimax squared-error lower bound contains a privacy term of order $\sigma^2d^2\log(1/\delta)/(n\varepsilon)^2$ for $d$-dimensional sub-Gaussian mean estimation, and of order $\sigma^2d^2/(n\varepsilon)^2$ for $d$-dimensional linear regression with bounded covariates and well-conditioned design covariance, where $\sigma^2$ is the noise variance proxy \citep{cai2021cost}. For sufficiently small $\epsilon_n$, $\epsilon_n$-GDP implies $(\varepsilon,\delta)$-DP with $\varepsilon$ of order $\epsilon_n\sqrt{\log(1/\delta)}$ \citep{dong2022gaussian}, so these lower bounds apply to such GDP mechanisms up to logarithmic factors. Our mechanism-error scale has the same scale on sample size and the privacy parameter, although the privacy guarantee and error criterion differ. The GDP guarantee in Theorem~\ref{thm: strong conv utility} follows from Proposition~\ref{prop: dp guarantee of exp mech}, which was established by \citet{gopi2022private}. The proof rescales the density around $\hat\theta_n$ and uses a second-order Taylor expansion to identify the Gaussian limit; strong convexity provides an integrable Gaussian bound, ensuring that the density is well-defined even on unbounded $\Theta$.

However, empirical loss functions are merely convex in many settings; for example, squared-error regression has a rank-deficient Hessian with a worst-case dataset. In such settings, we use the regularized objective in \eqref{eq: regularized empirical loss}. The following lemma characterizes the mechanism-error and regularization components of \eqref{eq: total error decomposition}.

\begin{restatable}[CLT for GDP RegExpMech under convex loss]{lemma}{restateLemRegularizedConvexUtility}\label{lem: regularized convex utility}
Suppose $F_n$ is convex over $\Theta$, $0<\mu_n=o(n)$, and (A1)--(A4) hold. If $\epsilon_n\sqrt{n\mu_n}\to\infty$, then RegExpMech in \eqref{equ: reg exp mech} satisfies
\begin{align*}
\epsilon_n\sqrt{n\mu_n}\,(\theta_n^{DP}-\theta_n^\#) &\dlim N_d(0,L^2H^{-1}),\\
\theta_n^\#-\hat\theta_n &=-\frac{\mu_n}{n}H^{-1}\theta_0+O\!\left(\frac{\mu_n}{n^{3/2-\zeta}}+\frac{\mu_n^2}{n^2}\right).
\end{align*}
\end{restatable}

Lemma~\ref{lem: regularized convex utility} characterizes the trade-off between mechanism-induced randomness and regularization bias. In particular, allowing the privacy level $\epsilon_n$ to vary with $n$, the DP noise is of order $1/(\epsilon_n\sqrt{n\mu_n})$, while the regularization bias is of order $\mu_n/n$. Thus, increasing the regularization parameter $\mu_n$ reduces the variability introduced by the exponential mechanism, but increases the bias due to regularization.


The ratio of the first-order bias scale to the mechanism-noise scale is
\begin{equation}\label{eq: joint tuning scales}
\kappa_n=\frac{\epsilon_n\mu_n^{3/2}}{\sqrt n}.
\end{equation}
The following theorem characterizes the asymptotic behavior according to whether $\kappa_n$ tends to zero, a positive constant, or infinity.

\begin{restatable}[Asymptotics of the GDP RegExpMech Error]{theorem}{restateThmRegularizedConvexUtility}\label{thm: regularized convex utility}
Under the assumptions of Lemma~\ref{lem: regularized convex utility}, and leting $\kappa_n$ be as in \eqref{eq: joint tuning scales}, we have
\begin{enumerate}
\item[(i)] \emph{Variance-dominated regime ($\kappa_n\to0$):} $\epsilon_n\sqrt{n\mu_n}\,(\theta_n^{DP}-\hat\theta_n)\dlim N_d(0,L^2H^{-1})$.

\item[(ii)] \emph{Balanced regime ($\kappa_n\to\kappa\in(0,\infty)$):} $\epsilon_n\sqrt{n\mu_n}\,(\theta_n^{DP}-\hat\theta_n)\dlim N_d(-\kappa H^{-1}\theta_0,L^2H^{-1})$.

\item[(iii)] \emph{Bias-dominated regime ($\kappa_n\to\infty$):} $\frac{n}{\mu_n}(\theta_n^{DP}-\hat\theta_n)\xrightarrow{P}-H^{-1}\theta_0$.
\end{enumerate}
\end{restatable}

Theorem \ref{thm: regularized convex utility} characterizes the trade-off between DP noise and regularization bias under GDP ExpMech, revealing a balanced regime through their convergence rates. More importantly, this finding reveals a fundamental distinction between the exponential mechanism under GDP and under $\varepsilon$-DP. Specifically, under $\varepsilon$-DP, the privacy-induced noise is not asymptotically negligible even for strongly convex losses \citep{awan2019benefits}. This positions the ExpMech as a practically and theoretically attractive approach to private estimation under GDP. 

The balanced regime in Theorem~\ref{thm: regularized convex utility} occurs when the mechanism noise and regularization bias have the same order. In particular, choosing $\mu_n=(n/\epsilon_n^2)^{1/3}$ with $\epsilon_n n\to\infty$ gives
\begin{equation}\label{eq: balanced privacy normalization}
(\epsilon_n n)^{2/3}(\theta_n^{DP}-\hat\theta_n)\dlim N_d(-H^{-1}\theta_0,L^2H^{-1}).
\end{equation}
For fixed $\epsilon_n=\epsilon>0$, this recovers a $n^{1/3}$ regularization scale and noise scale of $n^{-2/3}$. The following corollary formally specializes the three regimes to the case of a fixed privacy budget.

\begin{restatable}[Fixed-Privacy Specialization]{corollary}{restateCorBalancedTuning}\label{cor: balanced tuning}
Under the fixed-dataset regularity and convexity assumptions of Lemma~\ref{lem: regularized convex utility}, let $\epsilon_n=\epsilon>0$. Then, for
\begin{enumerate}
\item[(i)] $\mu_n=n^{2r-1}$ with $r\in(1/2,2/3)$,
$n^r(\theta_n^{DP}-\hat\theta_n)\dlim N_d(0,L^2H^{-1}/\epsilon^2)$.
\item[(ii)] $\mu_n=n^{1/3}$,
$n^{2/3}(\theta_n^{DP}-\hat\theta_n)\dlim N_d(-H^{-1}\theta_0,L^2H^{-1}/\epsilon^2)$.
\item[(iii)] $\mu_n=n^{1-r}$ with $r\in(0,2/3)$,
$n^r(\theta_n^{DP}-\hat\theta_n)\xrightarrow{P}-H^{-1}\theta_0$.
\end{enumerate}
\end{restatable}

Theorem~\ref{thm: regularized convex utility} and Corollary~\ref{cor: balanced tuning} show that regularization can introduce a nontrivial shift in the limiting distribution. The following corollary removes the first-order regularization bias and gives a centered Gaussian limit. In Section~\ref{sec: joint asymp}, we combine this conditional limit with the sampling distribution of the non-private estimator.

\begin{restatable}[Conditional First-Order Bias Correction]{corollary}{restateCorConditionalBiasCorrection}\label{cor: conditional bias correction}
Under the assumptions of Lemma~\ref{lem: regularized convex utility}, suppose
\begin{equation}\label{eq: conditional correction tuning}
\epsilon_n\mu_n^{3/2}=o(n^{1-\zeta}),\qquad \epsilon_n\mu_n^{5/2}=o(n^{3/2}).
\end{equation}
Let $\widetilde\theta_n=(I_d+\frac{\mu_n}{n}H^{-1})\theta_n^{DP}$. Then, $\epsilon_n\sqrt{n\mu_n}\,(\widetilde\theta_n-\hat\theta_n)\dlim N_d(0,L^2H^{-1})$. In particular, for fixed $\epsilon_n=\epsilon>0$ and $\mu_n=n^\gamma$ with $0<\gamma<3/5$, $n^{(1+\gamma)/2}(\widetilde\theta_n-\hat\theta_n)
\dlim N_d(0,L^2H^{-1}/\epsilon^2)$.
\end{restatable}

The balanced regime satisfies the correction conditions automatically under the assumptions of Lemma~\ref{lem: regularized convex utility}. The fixed-budget range above is sufficient; additional Hessian stability can permit larger regularization parameters, as shown in Corollary~\ref{cor: hessian stability}. The rate calculations are included in the proof.

\section{Joint Asymptotics and Private Confidence Regions} \label{sec: joint asymp}
In this section, we consider joint asymptotics that consider both the randomness in the data as well as from the privacy mechanism. 
We first establish a general joint convergence theorem, which we apply to construct private confidence regions based on the ExpMech.

\subsection{Joint Asymptotics on Data and Mechanism Randomness}

The exponential mechanism's distribution is specified conditional on the dataset, so its error $\theta_n^{DP}-\hat\theta_n$ and the sampling error $\hat\theta_n-\theta_0$ are generally dependent in finite samples. Even if their individual limiting distributions do not seem to depend on each other, this alone does not establish asymptotic independence, as the following example illustrates.

\begin{example}[Marginal convergence does not imply joint convergence] 
Let $Z\sim N(0,1)$ and define $X_n=Z$ and $Y_n=-Z$ for every $n$. Both sequences converge in distribution to $N(0,1)$, but it is not the case that $(X_n,Y_n) \overset d \rightarrow N(0,I)$; rather, we see that
\[
(X_n,Y_n)\dlim(Z,-Z).
\]
The limiting components have covariance $-1$ and are therefore dependent. Thus, the marginal limiting distributions cannot generally be combined without understanding the joint distributions. 
\end{example}

The following theorem establishes asymptotic independence by requiring a common conditional mechanism limit across a probability-one set of dataset sequences.

\begin{restatable}[Stacking Theorem for Joint Asymptotics]
{theorem}{restateThmJointConvergence}
\label{thm: joint convergence}
Let $(D_n)_{n\ge1}$ be a sequence of random datasets, with $D_n\in\mathcal S^n$, where $\mathcal S$ is Polish. Let $Q^\infty$ denote the joint law of this sequence and $Q^n$ the law of $D_n$. Write $\mathbf s_n$ for a fixed possible value of $D_n$.

Let $g_n:\mathcal S^n\to\mathbb R^m$ be measurable functions and $c\in\mathbb R^m$ a fixed vector such that
\[
G=\{(\mathbf s_n)_{n\ge1}:g_n(\mathbf s_n)\to c\},
\qquad Q^\infty(G)=1.
\]
Let $\hat\theta_n:\mathcal S^n\to\mathbb R^d$ be a measurable estimator. Given a dataset $\mathbf s_n$, the mechanism draws an output $\theta_n^{\mathrm{mech}}(\mathbf s_n)\in\mathbb R^d$ from a specified conditional distribution. Assume that the probability of every Borel output set is a measurable function of the dataset.

Suppose there exist fixed $a,b>0$ and random vectors $X,Y\in\mathbb R^d$ satisfying
\begin{enumerate}
\item[\textnormal{(B1)}] \emph{Sampling limit.}
Under the data randomness, $n^a(\hat\theta_n(D_n)-\theta_0)\dlim X$.

\item[\textnormal{(B2)}] \emph{Conditional mechanism limit.}
For every fixed sequence $(\mathbf s_n)_{n\ge1}\in G$,  we have $n^b\bigl(\theta_n^{\mathrm{mech}}(\mathbf s_n)
-\hat\theta_n(\mathbf s_n)\bigr)\dlim Y$, where convergence is over the mechanism randomness. The law of $Y$ may depend on $c$, but is otherwise the same for every sequence in $G$.
\end{enumerate}

Then, over both the data randomness governed by $Q^\infty$ and the mechanism randomness,
\[
\begin{pmatrix}
n^a(\hat\theta_n(D_n)-\theta_0)\\[2mm]
n^b(\theta_n^{\mathrm{mech}}(D_n)-\hat\theta_n(D_n))
\end{pmatrix}
\dlim
\begin{pmatrix}X\\Y\end{pmatrix},
\]
where the limiting vectors have the laws specified above and are independent.
\end{restatable}

\begin{proof}[Proof Sketch]
The proof couples the randomness from data sampling with the randomness from the privacy mechanism in three steps. First, the Skorohod representation theorem expresses the randomness in the data through a single uniform random variable while preserving the finite sample distribution at every $n$ as well as the limiting distribution. Second, since the mechanism acts conditionally on the realized dataset, we impute the data through its conditional distribution, which exists and is measurable by the Polish-space structure, and apply a second Skorohod representation to express the mechanism randomness through an independent uniform random variable. Combining the two representations yields joint convergence.  Because the limiting mechanism law depends on the dataset sequence only through the fixed vector $c$, its limit can be represented using only the second uniform random variable, and hence the two limits are independent.
\end{proof}

Theorem~\ref{thm: joint convergence} is a conditional-to-joint convergence principle related to those used in bootstrap asymptotics \citep{bucher2019conditional,wang2025differentially,awan2025one}. Here its role is to combine the conditional ExpMech limit with the M-estimator CLT. Note that independence between the two errors is \emph{not} required in finite samples. The almost-sure formulation of (B2) matches the fixed-dataset analysis of Section~\ref{sec: dp error asymp}; convergence in probability of the conditional laws would also suffice.

\subsection{Joint Asymptotic Normality with ExpMech}\label{sec: joint exp mech}

We apply Theorem~\ref{thm: joint convergence} to the ExpMech under the following population conditions.

Return to the i.i.d. sampling setup of Section~\ref{sec: m estimation setup}, and write $\Pi^\infty$ for the law of the resulting dataset sequence. Assume $F_n$ is convex over $\Theta$ almost surely for every $n$ and $\mathbb E|f_\theta(S)|<\infty$ for each $\theta\in\Theta$. In addition to (A1), we impose the following population conditions:
\begin{enumerate}
\item[\textbf{(A2$'$)}] \textbf{Population Smoothness and Integrability.}
For $\Pi$-almost every $s$, the local twice continuous differentiability and Hessian Lipschitz conditions in (A2) hold on a common neighborhood $\mathcal U$ of $\theta_0$, with a finite nonnegative measurable function $M$ satisfying $\mathbb E[M(S)]<\infty$. The loss-difference Lipschitz condition in (A2) holds for all $s,s'\in\mathcal S$ with a common finite constant $L$. In addition, the matrices $H$ and $\mathcal I$ defined in Section~\ref{sec: m estimation setup} exist and are finite, with $H\succ0$.

\item[\textbf{(A3$'$)}] \textbf{Hessian Second Moment.}
The Hessian at $\theta_0$ satisfies $\mathbb E\|\nabla^2f_{\theta_0}(S)\|^2<\infty$.

\item[\textbf{(A4$'$)}] \textbf{Data CLT.}
The non-private estimator satisfies $\sqrt n(\hat\theta_n-\theta_0)\dlim N_d(0,H^{-1}\mathcal I H^{-1})$.
\end{enumerate}

Under the i.i.d. sampling and convexity conditions above, (A2$'$) and (A3$'$) allow us to verify the fixed-dataset conditions (A2)--(A4) hold almost surely. In particular, the strong law gives the bounded sample averages required in (A2), while the second-moment conditions give the Hessian and minimizer rates. The following proposition establishes these properties. Assumption (A4$'$) then supplies the sampling limit that we combine with the conditional mechanism limit.

Let $G$ be the set of dataset sequences for which the assumptions of Section~\ref{sec: assumptions} hold; that is,
\begin{equation}\label{eq: almost sure data sequences}
G=\{(D_n)_{n\ge1}:\text{(A1)--(A4) hold}\}.
\end{equation}
Here $\zeta$ and $\mu_n$ are the same as in Section~\ref{sec: assumptions}. To connect with Theorem~\ref{thm: joint convergence}, take
\[
g_n(D_n)=\left(
\frac1n\sum_{i=1}^n M(s_i),\,
n^{1/2-\zeta}\|\nabla^2F_n(\theta_0)-H\|,\,
n^{1/2-\zeta}\|\hat\theta_n-\theta_0\|,\,
\|\theta_n^\#-\theta_0\|
\right),
\]
and $c=(\mathbb E[M(S)],0,0,0)$, omitting the last coordinate in the unregularized case. The following proposition establishes both $\Pi^\infty(G)=1$ and $g_n(D_n)\to c$ almost surely. Thus, Theorem~\ref{thm: joint convergence} applies on the probability-one subset of $G$ where this convergence holds.

\begin{restatable}[Almost-Sure Conditional Regularity with Rates]
{proposition}{restatePropAsRegularity}\label{prop: as regularity}
Under the sampling, convexity, and integrability conditions of Section~\ref{sec: joint exp mech}, together with (A1), (A2$'$), and (A3$'$), we have $\Pi^\infty(G)=1$ and $g_n(D_n)\to c$ almost surely.
\end{restatable}

Proposition~\ref{prop: as regularity} shows that the assumptions used in the conditional analysis hold almost surely under the population conditions without requiring (A4$'$). We now impose the data CLT (A4$'$) and apply Theorem~\ref{thm: joint convergence} to combine the sampling error with the conditional mechanism error. This gives the following joint asymptotic results.

\begin{restatable}[Joint Asymptotic Normality with GDP Exponential Mechanisms]{theorem}{restateThmExpMechJoint}
\label{thm: exp mech joint}
Under the sampling, convexity, and integrability conditions of Section~\ref{sec: joint exp mech}, together with (A1), (A2$'$), (A3$'$), and (A4$'$), use $\Sigma_1$ as defined in Section~\ref{sec: m estimation setup}.
\begin{enumerate}
\item \emph{Strongly convex loss.} For ExpMech in Theorem~\ref{thm: strong conv utility}, if $\epsilon_n n\to\infty$, then
\[
\begin{pmatrix}\sqrt n(\hat\theta_n-\theta_0)\\
\epsilon_n n(\theta_n^{DP}-\hat\theta_n)\end{pmatrix}
\dlim N_{2d}\!\left(0,\begin{pmatrix}\Sigma_1&0\\0&(L^2/\mu)H^{-1}\end{pmatrix}\right).
\]
\item \emph{Convex loss under balanced regularization.} For RegExpMech with $\kappa_n\to\kappa\in(0,\infty)$ defined in \eqref{eq: joint tuning scales}, suppose $0<\mu_n=o(n)$, and $\epsilon_n\sqrt{n\mu_n}\to\infty$, the corrected limit is
\[
\begin{pmatrix}\sqrt n(\hat\theta_n-\theta_0)\\
\epsilon_n\sqrt{n\mu_n}(\widetilde\theta_n-\hat\theta_n)\end{pmatrix}
\dlim N_{2d}\!\left(0,\begin{pmatrix}\Sigma_1&0\\0&L^2H^{-1}\end{pmatrix}\right),
\]
where $\widetilde\theta_n=(I_d+(\mu_n/n)H^{-1})\theta_n^{DP}$.
\end{enumerate}
\end{restatable}

The zero off-diagonal blocks establish asymptotic independence of sampling and mechanism errors. In the balanced regime for convex losses, no additional conditions on $\epsilon_n$ and $\mu_n$ are needed for the oracle bias correction. Theorem~\ref{thm: exp mech joint general} in Appendix~\ref{app: general inference} gives the complete joint result for all regimes.

\begin{restatable}[Extended Joint Asymptotics under Hessian Stability]{corollary}{restateCorHessianStability}\label{cor: hessian stability}
Under the population assumptions of Theorem~\ref{thm: exp mech joint}, suppose $F_n$ is convex, $0<\mu_n=o(n)$, and $\epsilon_n\sqrt{n\mu_n}\to\infty$. Let $H_n=\nabla^2F_n(\hat\theta_n)$ and define $\rho_n=\sup_{t\in[0,1]}\|\nabla^2F_n(\hat\theta_n+t(\theta_n^\#-\hat\theta_n))-H_n\|$. If $\epsilon_n\mu_n^{3/2}\rho_n=o_p(n^{1/2})$, then
\begin{align*}
\begin{pmatrix}
\sqrt n(\hat\theta_n-\theta_0)\\
\epsilon_n\sqrt{n\mu_n}\,(\widetilde\theta_n-\hat\theta_n)
\end{pmatrix}
\dlim N_{2d}\!\left(0,\begin{pmatrix}
H^{-1}\mathcal I H^{-1}&0\\
0&L^2H^{-1}
\end{pmatrix}\right),
\end{align*}
where $\widetilde\theta_n=(I_d+\frac{\mu_n}{n}H_n^{-1})\theta_n^{DP}$.
\end{restatable}

Under the specific loss structure---controlled Hessian variation between the two empirical minimizers---in Corollary~\ref{cor: hessian stability}, the first-order bias correction on the joint limit can be extended beyond the balanced regime. For quadratic losses the Hessian is constant, so $\rho_n=0$ and the stability condition holds automatically. Rate comparisons and the squared-error example are given with the proof.



\subsection{Asymptotic Confidence Region}\label{sec: confidence region}

We construct confidence regions that account for both sampling variability and mechanism noise. For RegExpMech, we use the balanced regime and a privately estimated first-order bias correction.

\begin{restatable}[Asymptotic Confidence Region]{theorem}{restateThmConfidenceRegion}\label{thm: confidence region}
Under the sampling, convexity, and integrability conditions of Section~\ref{sec: joint exp mech}, together with (A1), (A2$'$), (A3$'$), and (A4$'$), assume $H^{-1}\mathcal I H^{-1}\succ0$. Let $\hat\Sigma_{1,n}$ and $\hat\Sigma_{2,n}$ be private symmetric consistent estimators of $H^{-1}\mathcal I H^{-1}$ and $H^{-1}$, respectively, and let $\epsilon_n>0$ be deterministic.
\begin{enumerate}
\item \emph{Strongly convex loss:} For ExpMech in Theorem~\ref{thm: strong conv utility}, suppose $\epsilon_n n\to\infty$ and set
\[
\hat V_n=\hat\Sigma_{1,n}+\frac{L^2}{\mu\epsilon_n^2n}\hat\Sigma_{2,n},\qquad
\tilde\theta_n=\theta_n^{DP}.
\]
\item \emph{Convex loss with balanced regularization:} For RegExpMech, suppose $0<\mu_n=o(n)$, $\epsilon_n\sqrt{n\mu_n}\to\infty$, and $\kappa_n\to\kappa\in(0,\infty)$. Set
\[
\hat V_n=\hat\Sigma_{1,n}+\frac{L^2}{\epsilon_n^2\mu_n}\hat\Sigma_{2,n},\qquad
\tilde\theta_n=\left(I_d+\frac{\mu_n}{n}\hat\Sigma_{2,n}\right)\theta_n^{DP}.
\]
\end{enumerate}
Then $\sqrt n\,\hat V_n^{-1/2}(\tilde\theta_n-\theta_0)\dlim N_d(0,I_d)$, and
\[
\mathcal C_n(\hat V_n)=\left\{\theta\in\mathbb R^d:
n(\tilde\theta_n-\theta)^\top\hat V_n^{-1}(\tilde\theta_n-\theta)
\le\chi^2_{d,1-\alpha}\right\}
\]
is an asymptotic $(1-\alpha)$ confidence region for $\theta_0$.
\end{restatable}

Balanced regularzation makes consistency of the private covariance estimators sufficient for the estimated bias correction; no additional estimation rate is required. Theorem~\ref{thm: confidence region general} in Appendix~\ref{app: general inference} retains the general regularization conditions and both corrected and uncorrected centering options.

\begin{remark}
\label{rem: fixed privacy inference}
Under the assumptions of Theorem~\ref{thm: confidence region}, let $\epsilon_n=\epsilon>0$. In the strongly convex case, $\tilde\theta_n=\theta_n^{DP}$ and $\hat V_n=\hat\Sigma_{1,n}+\frac{L^2}{\mu\epsilon^2n}\hat\Sigma_{2,n}$. In the balanced regularized case, take $\mu_n=n^{1/3}$, so that $\tilde\theta_n=(I_d+n^{-2/3}\hat\Sigma_{2,n})\theta_n^{DP}$ and $\hat V_n=\hat\Sigma_{1,n}+\frac{L^2}{\epsilon^2n^{1/3}}\hat\Sigma_{2,n}$. In both cases, $\sqrt n\,\hat V_n^{-1/2}(\tilde\theta_n-\theta_0)\dlim N_d(0,I_d)$, and hence the confidence region $\mathcal C_n(\hat V_n)$ defined in Theorem~\ref{thm: confidence region} has asymptotic coverage $1-\alpha$.    
\end{remark}

Theorem~\ref{thm: confidence region} allows the privacy budget to vary with sample size. Remark~\ref{rem: fixed privacy inference} gives the fixed-budget construction used in the simulations, retaining both the regularization-bias correction and the mechanism-variance contribution.

\begin{algorithm}[ht!]
\caption{Private Confidence Region via the Exponential Mechanism}\label{alg: private CI}
\begin{algorithmic}[1]
\REQUIRE Loss $f$, dataset $D = \{s_i\}_{i=1}^n$, parameter space $\Theta_B$, confidence level $\alpha$, GDP budget $\epsilon_{\rm total}$, MCMC iterations $T$, burn-in $T_0$, constants $L, t, R, \eta$, and either exponent $\gamma \in (0,1)$ (convex) or strong convexity parameter $\mu > 0$ (strongly convex)
\STATE Set $\epsilon_{\rm main} = \epsilon_H = \epsilon_\fisher = \epsilon_{\rm total}/\sqrt{3}$ and $F_n(\theta) = n^{-1}\sum_{i=1}^n f(\theta; s_i)$
\IF{convex case}
    \STATE Set $Q(\theta) = F_n(\theta) + \frac{n^\gamma}{2n}\|\theta\|_2^2$ and $k = \epsilon_{\rm main}^2 n^{1+\gamma}/L^2$
\ELSE[ strongly convex case ]
    \STATE Set $Q(\theta) = F_n(\theta)$ and $k = \epsilon_{\rm main}^2 \mu n^2/L^2$
\ENDIF
\STATE Draw $\theta_n^{\rm DP} \leftarrow \mathsf{RWMH}(Q, k, \Theta_B, T, T_0)$
\STATE $(\hat\Sigma_1, \hat\Sigma_2) \leftarrow \mathsf{PrivCov}(f, \theta_n^{\rm DP}, D, \epsilon_H, \epsilon_\fisher, t, R, \eta)$
\IF{convex case}
    \STATE Set $\hat V_n = \hat\Sigma_1 + (L^2/\epsilon_{\rm main}^2) \cdot n^{-\gamma}\,\hat\Sigma_2$
    \STATE If $\gamma \ge 1/3$, set $\tilde\theta_n = (I_d + n^{-(1-\gamma)}\hat\Sigma_2)\,\theta_n^{\rm DP}$; otherwise $\tilde\theta_n = \theta_n^{\rm DP}$
\ELSE[ strongly convex case ]
    \STATE Set $\hat V_n = \hat\Sigma_1 + (L^2/(\epsilon_{\rm main}^2 \mu n))\,\hat\Sigma_2$ and $\tilde\theta_n = \theta_n^{\rm DP}$
\ENDIF
\ENSURE $(1 - \alpha)$ confidence region $\mathcal{C}_n = \bigl\{\theta : n(\tilde\theta_n - \theta)^\top \hat V_n^{-1}(\tilde\theta_n - \theta) \le \chi^2_{d, 1-\alpha}\bigr\}$ with $\epsilon_{\rm total}$-GDP guarantee
\end{algorithmic}
\end{algorithm}

Algorithm~\ref{alg: private CI} implements the displayed confidence-region formula. Its theoretical validity requires the assumptions of Theorem~\ref{thm: confidence region}, including consistency of the private covariance estimates. The total GDP budget is split equally across three releases: the mechanism draw, the private Hessian, and the private Fisher information matrix. This equal allocation is a simple default; optimizing the split for finite-sample performance is an open direction. The private estimator $\theta_n^{\rm DP}$ is obtained by sampling from the ExpMech density via $\mathsf{RWMH}$ (Algorithm~\ref{alg: M-H sampling for ExpMech}), after which $\mathsf{PrivCov}$ (Algorithm~\ref{alg: private covariance}) is invoked at $\theta_n^{\rm DP}$ to produce the private covariance estimates $\hat\Sigma_1$ and $\hat\Sigma_2$ for the covariance matrices $H^{-1}\fisher H^{-1}$ and $H^{-1}$, respectively. The variance estimator $\hat V_n$ retains the leading-order contributions from both sampling variability $\hat\Sigma_1$ and privacy noise $\hat\Sigma_2$, ensuring that the confidence region width reflects the actual privacy cost rather than only the statistical uncertainty. In the convex case, the main-text construction uses balanced tuning and the corrected center. At a fixed privacy parameter, the simulations use $\mu_n=n^{1/3}$ as in Remark~\ref{rem: fixed privacy inference}; the general centering options are retained in Theorem~\ref{thm: confidence region general}. In the strongly convex case, no regularization is imposed and no bias correction is needed.

\section{Simulation}\label{sec: simulation}

In this section, we illustrate the finite-sample performance of our confidence regions through simulation studies on logistic and linear regression, validating our asymptotic theory and comparing coverage and effective diameter against existing DP uncertainty quantifying methods as baselines.

We use logistic regression as the primary model. The data-generating process draws $X_i \sim \mathrm{Uniform}([-1,1]^d)$ independently and $Y_i \mid X_i \sim \mathrm{Bernoulli}(\sigma(X_i^\top \theta_0))$, where $\sigma(z) = (1 + e^{-z})^{-1}$ is the sigmoid function. The feature bound implied by the uniform box design is $\|x_i\|_2 \le \sqrt{d}$. Linear regression is included as a secondary model in Section~\ref{subsec: convergence}, with $Y_i = X_i^\top\theta_0 + \xi_i$, $\xi_i \overset{\rm iid}{\sim} \mathcal{N}(0,1)$. Responses are clipped to $[-1,1]$ prior to privatization to ensure bounded sensitivity.

\textbf{Simulation Setup:} We fix $d = 6,\, B = 2,\, \theta_0 = (1,\,-1,\,0.5,\,-0.5,\,0.25,\,-0.25)^\top$, where the parameter space is the $\ell_2$ ball $\Theta_B = \{\theta : \|\theta\|_2 \le B\}$. For logistic regression, the difference-Lipschitz constant is $L = 2\sqrt{d}$ and the Hessian spectral bound is $t = d/4$. For linear regression, $L = 2\sqrt{d}\,(\sqrt{d}\,B + 1)$ and $t = d$~(see Example~\ref{eg: logistic_LM_Lipshitz}). The sample size $n$ is run across $\{200, 400, 800, 1600, 3200, 6400, 12800\}$.

For our method, it is an approximate sampling from the ExpMech density carried out via Random-Walk Metropolis-Hastings, shorten as $\mathsf{RWMH}$ (Algorithm~\ref{alg: M-H sampling for ExpMech} in Appendix~\ref{appendix: algorithms}), run for $2{,}000$ iterations with the first $1{,}000$ discarded as burn-in.

The total GDP privacy budget is $\epsilon_{\rm total} = 0.6$, split equally as $\epsilon_{\rm main} = \epsilon_H = \epsilon_S = \epsilon_{\rm total}/\sqrt{3}$ when there are mechanism draw, private Hessian release, and private score covariance release respectively. The $\varepsilon$-DP baselines use $0.4807$-DP, roughly comparable to $0.6$-GDP under the trade-off function matching in Section~\ref{appendix: more background} \citep{awan2024optimizing}. The nominal coverage level is set to be 0.95 and $R = 1000$ replications throughout. Simulations were conducted on a computing cluster. Replicates were parallelized across cores, with each sample size run as a separate single-node job on 32 to 64 CPU cores in computationally heavy setting. Detailed descriptions of all algorithms and competing methods are given in Appendix~\ref{appendix: more background}.

\textbf{Mechanisms.}
Methods used in the entire section includes (1) the GDP ExpMech (implemented by Algorithm~\ref{alg: M-H sampling for ExpMech}), (2) differentially private noisy gradient descent (DP-NGD)~\citep{avella2023differentially} with GDP guarantee,  (3) objective perturbation ~\citep{wang2018differentially} with $\varepsilon$-DP guarantee, and (4) DP-GD and DP-SGD~\citep{abadi2016deep} with CI approach from \citet{xia2025inference} and with asymptotic GDP guarantee. The DP-NGD we implemented from \citet{avella2023differentially} is their adjusted version that the private sandwich covariance with a correction for the variance accumulated over $K$ noisy gradient steps. 
Not all methods appear in every task; we specify the relevant subset in each subsection.

\subsection{Validation of Asymptotic Convergence}\label{subsec: convergence}

To demonstrate our private estimator converges to the non-private estimator indicated by Theorem~\ref{thm: regularized convex utility} with $\gamma \in \{0.4,\, 0.3\Bar{3},\, 0.2\}$, we validate it by comparing the Mahalanobis statistic to a chi-square distribution
\begin{align*}
  \mathbf{z}_n =
  \begin{cases}
    n^{(1+\gamma)/2}\!\left(\theta_n^{\rm DP} - \hat\theta_n\right), 
      & 0 < \gamma < 1/3, \\[4pt]
    n^{2/3}\!\left(\theta_n^{\rm DP} - \hat\theta_n\right) + H_n(\hat\theta_n)^{-1}\hat\theta_n, 
      & \gamma = 1/3, \\[4pt]
    (n\mu_n)^{1/2}\!\left(\widetilde{\theta}_n - \hat\theta_n\right),
      & 1/3 < \gamma < 3/5,
  \end{cases}
\end{align*}
where $\widetilde{\theta}_n = (I_d + \frac{\mu_n}{n}H_n(\hat\theta_n)^{-1})\theta_n^{\rm DP}$ and $H_n(\hat\theta_n) = \nabla^2 F_n(\hat\theta_n)$ is the non-private empirical Hessian evaluated at the non-private ERM, used as a consistent estimator of the population quantity $H(\theta_0)$ for diagnostic centering only. Since this subsection is dedicated to verifying the CLT behavior from the privacy noise, no private release of $H_n(\hat\theta_n)$ is performed in this task. The simulation is run for both logistic and linear regression. Agreement between the empirical law of $T_n$ and $\chi^2_d$ is quantified by the $1$-Wasserstein distance $\mathcal{W}_1$ and the Kolmogorov--Smirnov statistic (smaller is better).

\begin{table}[ht]
\centering
\resizebox{\textwidth}{!}{%
\begin{tabular}{rl rr rr rr}
\toprule
& & \multicolumn{2}{c}{$\gamma=0.40$} & \multicolumn{2}{c}{$\gamma=0.3\Bar{3}$} & \multicolumn{2}{c}{$\gamma=0.20$} \\
\cmidrule(lr){3-4}\cmidrule(lr){5-6}\cmidrule(lr){7-8}
Method & $n$ & $W$ & KS & $W$ & KS & $W$ & KS \\
\midrule
 & 200 & 1.4320 (0.0164) & 0.1825 (0.0152) & 3.5677 (0.0723) & 0.5699 (0.0100) & 3.7840 (0.0773) & 0.6188 (0.0094) \\
 & 400 & 1.5262 (0.0243) & 0.1589 (0.0156) & 2.2795 (0.0474) & 0.3261 (0.0130) & 2.4918 (0.0520) & 0.3521 (0.0125) \\
 & 800 & 1.1608 (0.0215) & 0.1337 (0.0153) & 1.3302 (0.0233) & 0.1796 (0.0150) & 1.4903 (0.0299) & 0.1929 (0.0146) \\
Logistic & 1600 & 0.8028 (0.0132) & 0.1045 (0.0158) & 0.7302 (0.0202) & 0.1008 (0.0140) & 1.0020 (0.0169) & 0.1280 (0.0155) \\
 & 3200 & 0.4662 (0.0094) & 0.0633 (0.0153) & 0.5375 (0.0129) & 0.0679 (0.0150) & 0.5413 (0.0091) & 0.0792 (0.0154) \\
 & 6400 & 0.5367 (0.0101) & 0.0769 (0.0154) & 0.2520 (0.0069) & 0.0541 (0.0157) & 0.1394 (0.0082) & 0.0306 (0.0125) \\
 & 12800 & 0.1730 (0.0071) & 0.0295 (0.0101) & 0.1676 (0.0052) & 0.0295 (0.0122) & 0.1088 (0.0034) & 0.0194 (0.0081) \\
\midrule
 & 200 & 4.9824 (0.0966) & 0.9081 (0.0024) & 5.4380 (0.1025) & 0.9727 (0.0023) & 5.6835 (0.1056) & 0.9953 (0.0003) \\
 & 400 & 3.7564 (0.0797) & 0.6402 (0.0086) & 4.6161 (0.0923) & 0.8486 (0.0042) & 5.3132 (0.1010) & 0.9631 (0.0020) \\
 & 800 & 1.8045 (0.0446) & 0.2528 (0.0136) & 3.1577 (0.0673) & 0.4890 (0.0117) & 4.5082 (0.0899) & 0.8121 (0.0052) \\
Linear & 1600 & 0.1924 (0.0134) & 0.0282 (0.0061) & 1.3951 (0.0346) & 0.1826 (0.0134) & 3.1379 (0.0686) & 0.5039 (0.0097) \\
 & 3200 & 0.1993 (0.0035) & 0.0451 (0.0156) & 0.2703 (0.0107) & 0.0407 (0.0129) & 1.5826 (0.0384) & 0.2191 (0.0145) \\
 & 6400 & 0.1226 (0.0034) & 0.0249 (0.0145) & 0.1452 (0.0053) & 0.0329 (0.0150) & 0.2597 (0.0123) & 0.0411 (0.0128) \\
 & 12800 & 0.1162 (0.0056) & 0.0250 (0.0156) & 0.1271 (0.0063) & 0.0191 (0.0154) & 0.1693 (0.0041) & 0.0387 (0.0155) \\
\bottomrule
\end{tabular}
}
\caption{Convergence of the GDP ExpMech for logistic and linear regression, measured in 1-Wasserstein ($W$) and Kolmogorov-Smirnov (KS) statistics averaged over 1000 replicates with MCSEs.}
\label{tab:thm10-convergence}
\end{table}

In Table~\ref{tab:thm10-convergence}, the statistic for $\gamma=0.2$ uses the raw ExpMech output, with no bias correction. The statistics for $\gamma=1/3$ and $\gamma=0.4$ use the corrections displayed above, which are based on the non-private Hessian. Although $\gamma = 1/3$ induces stronger regularization, the DP noise and regularization bias converge at the same rate. Consequently, after correcting for the center of the Gaussian limiting distribution, the resulting convergence rate is $n^{2/3}$. This is faster than that of $\gamma = 0.2$, where no center correction is required, but its overall convergence rate is limited by the DP noise. Nevertheless, Table~\ref{tab:thm10-convergence} shows that $T_n$ is already very close to a chi-squared distribution with 6 degrees of freedom with respect to each of the metric when $n \ge 3200$.

\begin{table}[t]
\centering
\resizebox{0.45\textwidth}{!}{%
\begin{tabular}{rl rr}
\toprule
& & \multicolumn{2}{c}{$\gamma=0.8$} \\
\cmidrule(lr){3-4}
Method & $n$ & $W$ & KS \\
\midrule
 & 200   & $6.0672\;(0.0936)$ & $0.4707\;(0.0139)$ \\
 & 400   & $5.5937\;(0.1008)$ & $0.4222\;(0.0150)$ \\
 & 800   & $4.7901\;(0.0930)$ & $0.3850\;(0.0154)$ \\
Linear & 1600  & $3.8885\;(0.0756)$ & $0.3197\;(0.0149)$ \\
 & 3200  & $3.6642\;(0.0660)$ & $0.3278\;(0.0155)$ \\
 & 6400  & $3.1765\;(0.0575)$ & $0.2891\;(0.0158)$ \\
 & 12800 & $2.7028\;(0.0559)$ & $0.2320\;(0.0154)$ \\
\bottomrule
\end{tabular}%
}
\caption{Convergence of our ExpMech for linear regression with large bias regularization, measured in 1-Wasserstein ($W$) and Kolmogorov-Smirnov (KS) statistics averaged over 1000 replicates. MCSEs are reported in parentheses.}
\label{tab:cor-convergence}
\end{table}

Table~\ref{tab:cor-convergence} considers $\gamma=0.8$ for linear regression, using the same statistic with $\widetilde\theta_n$. This choice lies outside the range $\mu_n=o(n^{3/5})$ of Corollary~\ref{cor: conditional bias correction}, but Corollary~\ref{cor: hessian stability} applies because the Hessian of the squared-error loss is constant. As $n$ increases from $200$ to $12800$, the Wasserstein distance decreases from $6.07$ to $2.70$ and the KS statistic decreases from $0.47$ to $0.23$. These decreases agree with the corollary, although the chi-squared approximation remains less accurate than in Table~\ref{tab:thm10-convergence} at the sample sizes considered.

\subsection{Confidence Region Comparison}\label{subsec: CI comparison}

This subsection compares the uncertainty quantification of our method against other DP confidence region approaches. It is demonstrated through logistic regression model with nominal coverage set to $95\%$. For each method we report (i)~\emph{coverage} and (ii)~\emph{effective diameter}, which is defined as the diameter of the Euclidean ball with the same volume as $\mathcal{C}_n(\hat V_n)\cap\Theta_B$. Without the intersection, this quantity equals the geometric mean of the diameters of the ellipsoid along the eigenvectors of $\hat V_n$, namely $2(\chi^2_{d,1-\alpha}/n)^{1/2}(\det\hat V_n)^{1/(2d)}$. We compare our method (implemented by Algorithm~\ref{alg: private CI}) with $\gamma =1/3$, against those of \citet{avella2023differentially}, \citet{wang2018differentially}, \citet{xia2025inference} along with the naive baseline. The naive region uses only $\hat\Sigma_{1,n}$, centered at $\theta_n^{\rm DP}$, which is asymptotically valid, but fails to account for the uncertainty introduced by the privacy mechanism.

\begin{table}[t]
\centering
\setlength{\tabcolsep}{6pt}
\resizebox{0.75\textwidth}{!}{%
\begin{tabular}{rrrrrrr}
\toprule
$n$ & naive & Ours $\gamma=0.3\Bar{3}$ & Xia & Avella & Wang \\
\midrule
200   & 0.029 [1.786] & 1.000 [3.996] & 1.000 [3.968] & 0.678 [2.857] & 0.997 [3.967] \\
400   & 0.007 [1.310] & 0.995 [3.916] & 1.000 [3.999] & 0.579 [1.803] & 0.997 [3.993] \\
800   & 0.007 [0.940] & 0.982 [3.420] & 1.000 [3.999] & 0.655 [1.154] & 1.000 [3.929] \\
1600  & 0.012 [0.673] & 0.967 [2.477] & 1.000 [4.000] & 0.736 [0.762] & 0.986 [3.529] \\
3200  & 0.012 [0.478] & 0.950 [1.682] & 1.000 [4.000] & 0.842 [0.514] & 0.960 [2.171] \\
6400  & 0.024 [0.339] & 0.962 [1.121] & --            & 0.892 [0.352] & 0.923 [1.999] \\
12800 & 0.035 [0.241] & 0.942 [0.738] & --            & 0.929 [0.245] & 0.935 [0.648] \\
\bottomrule
\end{tabular}%
}
\caption{Coverage and confidence interval length for logistic regression with nominal coverage $95\%$, across state-of-the-art uncertainty quantification methods for different DP mechanisms. Each cell reports coverage with the confidence interval length in square brackets.
.}
\label{tab:CI-cov-leng}
\end{table}

Since \citet{wang2018differentially} provide only coordinate-wise confidence intervals, we construct a joint $95\%$ confidence region by calibrating the empirical Mahalanobis ellipsoid from the Monte Carlo draws of their algorithm. This is more efficient than a Bonferroni correction and makes no distributional assumptions. Due to the high computational cost of the CI procedure in \citet{xia2025inference}, it is run separately on the computing cluster for $n \in \{200, 400, 800, 1600, 3200\}$. 


Table~\ref{tab:CI-cov-leng} shows that our method with $\gamma=1/3$ is initially conservative but rapidly approaches the nominal $95\%$ coverage, reaching it by $n=3200$ while the effective diameter decreases substantially with increasing sample size. In contrast, \citet{xia2025inference} remains overly conservative, whereas the correction of \citet{avella2023differentially} persistently undercovers. Although the confidence regions based on \citet{wang2018differentially} come close to the nominal level in this low-dimensional setting, they exhibit slight undercoverage in large samples. Furthermore, the method of \citet{wang2018differentially} satisfies pure-DP, which is expected to have inferior scaling with dimension, motivating the higher-dimensional comparison in Table~\ref{tab:CI-dim-scaling}.

\begin{table}[t]
\centering
\setlength{\tabcolsep}{6pt}
\resizebox{0.9\textwidth}{!}{%
\begin{tabular}{rc cc cc cc}
\toprule
 & \multicolumn{2}{c}{$d = 18$} & \multicolumn{2}{c}{$d = 30$} & \multicolumn{2}{c}{$d = 42$} \\
\cmidrule(lr){2-3} \cmidrule(lr){4-5} \cmidrule(lr){6-7}
$n$ & Ours & Wang & Ours & Wang & Ours & Wang \\
\midrule
2000  & 1.000 [4.000] & 0.958 [4.000] & 1.000 [4.000] & 0.861 [4.000] & 1.000 [4.000] & 0.651 [4.000] \\
4000  & 0.991 [3.354] & 0.944 [3.640] & 1.000 [4.000] & 0.840 [4.000] & 1.000 [4.000] & 0.617 [4.000] \\
8000  & 0.962 [2.349] & 0.952 [2.236] & 0.993 [3.401] & 0.859 [3.803] & 1.000 [4.000] & 0.669 [4.000] \\
16000 & 0.941 [1.566] & 0.919 [1.178] & 0.954 [2.372] & 0.904 [1.823] & 0.991 [3.080] & 0.793 [2.573] \\
32000 & 0.941 [1.012] & 0.926 [0.616] & 0.942 [1.569] & 0.925 [0.927] & 0.949 [2.084] & 0.870 [1.252] \\
\bottomrule
\end{tabular}%
}
\caption{Coverage and confidence interval length for logistic regression across dimensions $d \in \{18, 30, 42\}$ with nominal coverage $95\%$. Each cell reports coverage with the confidence interval length in square brackets.
.}
\label{tab:CI-dim-scaling}
\end{table}

Table~\ref{tab:CI-dim-scaling} examines the effect of increasing dimension. For higher dimensions, we run the two methods for $n \in \{2000, 4000, 8000, 16000, 32000\}$. As the dimension grows, our method continues to achieve coverage close to the nominal $95\%$ level, whereas the confidence regions based on \citet{wang2018differentially} exhibit progressively worse undercoverage despite having comparable effective diameters. This behavior is consistent with the increasing regularization bias introduced by their procedure, which becomes more pronounced in higher dimensions, while our finite-sample correction continues to account for both sampling and privacy uncertainty.




\subsection{Real Data Analysis} \label{sec: real data}

The real data analysis section applies the procedures of Theorem~\ref{thm: strong conv utility} and Theorem~\ref{thm: confidence region} 
to a Huber regression on the California Housing dataset \citep{pace1997sparse}, which is publicly available via \texttt{sklearn}\footnote{\url{https://scikit-learn.org/stable/modules/generated/sklearn.datasets.fetch_california_housing.html}} and contains $n_{\rm full} = 20{,}640$ observations. We select four features ($x$): median income (MedInc), average number of rooms (AveRooms), latitude, and longitude ($d = 4$), which are significant factors of median house value, the response variable ($y$). Features and response are clipped at the 1st and 99th percentiles to remove outliers, then scaled to $[-1,1]$.


We use Huber regression on the four California Housing features, with a fixed ridge penalty $\rho=0.1$: 
$$
F_n^\rho(\theta) = n^{-1}\sum_i f_\delta(\theta;\, x_i, y_i) + (\rho/2)\|\theta\|_2^2,
$$ 
where 
\begin{align*}
  f_\delta(\theta;\, x_i, y_i) \;=\; 
  \begin{cases} 
  \tfrac{1}{2}(y_i - x_i^\top\theta)^2, & |y_i - x_i^\top\theta| \le \delta, \\ \delta\bigl(|y_i - x_i^\top\theta| - \tfrac{\delta}{2}\bigr), & \text{otherwise,} \end{cases}    
\end{align*}
Unlike the regularizer from RegExpMech, $\rho$ is fixed and non-vanishing, making $F_n^\rho$ $\rho$-strongly convex. The Huber loss is continuously differentiable, but its second derivative is discontinuous where $|y_i-x_i^\top\theta|=\delta$, so the smoothness condition in (A2) does not hold exactly. We use this loss intentionally to examine how our method performs when this condition is not fully met. The privacy guarantee is unaffected, because Proposition~\ref{prop: dp guarantee of exp mech} requires only strong convexity and the Lipschitz condition on loss differences, while the asymptotic results of Theorem~\ref{thm: strong conv utility} and Theorem~\ref{thm: confidence region} are not directly guaranteed. Note that the ground truth 
$$
\theta_0^\rho = \arg\min_\theta\{F(\theta)+(\rho/2)\|\theta\|_2^2\}
$$ 
is approximated by fitting $F_n^\rho$ on the full dataset. The in-sample estimator, denoted as $\hat\theta_n^\rho = \arg\min_\theta F_n^\rho(\theta)$, is fit on bootstrapping subsamples (drawn with replacement). On each subsample we run the ExpMech and \citet{avella2023differentially}'s CI and construct $\mathcal{C}_n(\hat V_n)$. Our method samples $\theta_n^{\rm DP}\propto\exp\{-k_{\rm sc}F_n^\rho(\theta)\}$ with $k_{\rm sc}=\epsilon_{\rm main}^2\rho n^2/L^2$, $\rho=0.1$, and $\delta=1$, using Algorithm~\ref{alg: M-H sampling for ExpMech}. The DP-NGD runs $K=100$ full-batch noisy gradient steps on $F_n^\rho$, warm-started from $\hat\theta_n^\rho$.

We compare with \citet{avella2023differentially} because both methods provide exact $\epsilon$-GDP guarantees and their theoretical utility results apply to ridge-penalized Huber regression. Our privacy calibration follows Proposition~\ref{prop: dp guarantee of exp mech} directly: $F_n^\rho$ is $\rho$-strongly convex by construction, and the ridge term, being data-independent, cancels in $f_{\theta}(s_i)-f_{\theta}(s_j)$, leaving $L=2\delta\sqrt d=4$ unchanged from the unpenalized loss. The convergence analysis of \citet{avella2023differentially} requires only local strong convexity, which is likewise guaranteed by the ridge penalty.

\begin{table}[t]
\centering
\resizebox{0.7\textwidth}{!}{%
\begin{tabular}{r cc @{\hspace{1.5em}} cc}
\toprule
& \multicolumn{2}{c}{Ours (RegExpMech)} & \multicolumn{2}{c}{Avella} \\
\cmidrule(lr){2-3} \cmidrule(lr){4-5}
$n$ & Coverage & Length & Coverage & Length \\
\midrule
 200 & 0.916 (0.0088) & 2.0453 (0.0052) & 0.890 (0.0099) & 2.4450 (0.0042) \\
 400 & 0.949 (0.0070) & 1.0648 (0.0018) & 0.887 (0.0100) & 1.2751 (0.0002) \\
 800 & 0.949 (0.0070) & 0.5395 (0.0006) & 0.899 (0.0095) & 0.6400 (0.0001) \\
1600 & 0.959 (0.0063) & 0.2728 (0.0002) & 0.886 (0.0100) & 0.3234 (0.0001) \\
3200 & 0.946 (0.0071) & 0.1408 (0.0001) & 0.923 (0.0084) & 0.1658 (0.0001) \\
6400 & 0.948 (0.0070) & 0.0749 (0.0001) & 0.908 (0.0091) & 0.0869 (0.0001) \\
12800 & 0.944 (0.0073) & 0.0749 (0.0001) & 0.921 (0.0085) & 0.0869 (0.0001) \\
\bottomrule
\end{tabular}
}%
\caption{Confidence interval lengths and coverage for strongly convex Huber ridge regression on the California Housing dataset, comparing our RegExpMech against DP-NGD (Adjusted) \citep{avella2023differentially}. MCSEs in parentheses.}
\label{tab:realData}
\end{table}




In Table~\ref{tab:realData} we see that our method approximately achieves the nominal $95\%$ coverage with sample sizes as small as $400$, converging faster than in the convex loss setting, whereas \citet{avella2023differentially} fails to attain nominal coverage in any of the sample sizes considered. This suggests that our confidence region can remain valid even when the smoothness condition in (A2) is not satisfied.

\section{Conclusion and Discussion} \label{sec: discussion}

We develop a joint asymptotic framework for inference with the Gaussian differentially private exponential mechanism. We establish asymptotic independence of sampling variability and mechanism-induced randomness, characterize when private estimation recovers the non-private limiting distribution, and quantify the tradeoff between privacy noise and regularization bias. These results yield confidence regions for the true parameter that account for both sources of uncertainty. Simulations and a real-data analysis show improved finite-sample coverage with competitive interval lengths. The joint convergence framework may be of independent interest, potentially applying to other post-data randomization procedures.

Several limitations of the current framework are worth noting. The confidence region in Theorem~\ref{thm: confidence region} requires a three-way privacy budget split among the primary mechanism draw, the private Hessian release, and the private score covariance release. While equal splitting performs well empirically, it is not provably optimal, and the best allocation depends on problem-specific quantities such as the condition number of $H$ and the signal-to-noise ratio. 

Our theory assumes exact sampling from the exponential mechanism, whereas the simulations use a finite-length Metropolis--Hastings sampler that only approximates the target distribution. Extending our results to approximate MCMC implementations with rigorous privacy guarantees is an important direction for future work \citep{lin2024tractable}. 


Finally, while the current asymptotic framework is developed in the fixed-dimensional setting, extending the theory to regimes where $d$ grows with $n$ raises fundamental questions about the interplay among dimensionality, privacy budget, and regularization strength that lie beyond the scope of the present work.

Several directions for future work follow naturally. Deriving data-adaptive privacy budget allocation rules under GDP, optimized with respect to estimable problem characteristics, could improve finite-sample performance without sacrificing privacy. Extending the joint asymptotic independence result of Lemma~\ref{thm: joint convergence} to other mechanism classes, such as the $K$-norm gradient mechanism \citep{reimherr2019kng} or output perturbation \citep{wang2018differentially}, would further unify the theory of private inference. Finally, extending the framework to high-dimensional settings by leveraging sparsity or low-rank structure, potentially through privacy-aware dimensionality reduction as a preprocessing step, represents a promising direction for making the methodology applicable to modern large-scale problems.

\acks{The authors gratefully acknowledge support from NSF SES-2150615 and SES-2610910.}

\newpage
\appendix
\phantomsection\label{supplementary-material}
\bigskip

\begin{center}

{\large\bf SUPPLEMENTARY MATERIAL}

\end{center}

\section{Additional Simulation Results} \label{appendix: more simulation}

\subsection{Validation of Asymptotic Convergence (Section~\ref{subsec: convergence} Continued)}

\begin{table}[ht]
\centering
\resizebox{0.4\textwidth}{!}{%
\begin{tabular}{rr rr}
\toprule
& & \multicolumn{1}{c}{Logistic} & \multicolumn{1}{c}{Linear} \\
\cmidrule(lr){3-3}\cmidrule(lr){4-4}
$\gamma$ & $n$ & $\chi^2_6$ & $\chi^2_6$ \\
\midrule
1/3 &   200 & 2.432 (0.038) & 0.561 (0.007) \\
1/3 &   400 & 3.720 (0.062) & 1.383 (0.018) \\
1/3 &   800 & 4.669 (0.086) & 2.842 (0.043) \\
1/3 &  1600 & 5.272 (0.090) & 4.604 (0.076) \\
1/3 &  3200 & 5.462 (0.098) & 5.739 (0.100) \\
1/3 &  6400 & 5.752 (0.105) & 5.876 (0.105) \\
1/3 & 12800 & 5.969 (0.104) & 5.879 (0.105) \\
\midrule
0.20 &   200 & 2.215 (0.033) & 0.316 (0.004) \\
0.20 &   400 & 3.507 (0.058) & 0.686 (0.009) \\
0.20 &   800 & 4.509 (0.080) & 1.491 (0.020) \\
0.20 &  1600 & 4.997 (0.093) & 2.861 (0.041) \\
0.20 &  3200 & 5.458 (0.103) & 4.418 (0.071) \\
0.20 &  6400 & 5.866 (0.103) & 5.755 (0.098) \\
0.20 & 12800 & 5.931 (0.109) & 5.842 (0.109) \\
\bottomrule
\end{tabular}
}
\caption{Chi-squared statistic $\chi^2_6$ for logistic and linear regression, averaged over 1000 replicates with MCSEs.}
\label{tab:thm10-chisq}
\end{table}

Table~\ref{tab:thm10-chisq} shows that the average $\chi^2_6$ statistic converges to its theoretical mean of $6$ as the sample size increases, confirming the accuracy of the asymptotic approximation. Convergence is faster for $\gamma = 1/3$ than for $\gamma = 0.2$, which aligns with the finding of Theorem~\ref{thm: regularized convex utility}.

\subsection{Privacy Budget Sensitivity}\label{subsec: eps sensitivity}

This subsection illustrates the effect of privacy budgets on CI coverage, targeting $95\%$ coverage, and length by fixing sample size under logistic regression. We run our CI method on exponential mechanism by fixing $n = 3200$ and varying $\epsilon_{\rm total} \in \{0.2, 0.4, 0.6, 0.8, 1.0\}$. 


\begin{table}[t]
\centering
\resizebox{0.4\textwidth}{!}{%
\begin{tabular}{rrr}
\toprule
$\epsilon_{\mathrm{total}}$ & Coverage & CI length \\
\midrule
0.2 & 0.998 (0.0014) & 5.4245 (0.00500) \\
0.4 & 0.967 (0.0056) & 2.7123 (0.00127) \\
0.6 & 0.950 (0.0069) & 1.8423 (0.00077) \\
0.8 & 0.948 (0.0070) & 1.4157 (0.00050) \\
1.0 & 0.937 (0.0077) & 1.1691 (0.00040) \\
\bottomrule
\end{tabular}
}
\caption{Coverage and confidence interval lengths with MCSEs for logistic regression by our ExpMech CI method, varying $\epsilon_{\mathrm{total}}$.}
\label{tab:coverage-length_varying_budgets}
\end{table}

Table~\ref{tab:coverage-length_varying_budgets} shows that $\hat V_n$ appropriately adapts to the privacy budget once it is sufficiently large. This behavior is consistent with Theorem~\ref{thm: confidence region}. Specifically, with sample size $n=3200$ and the total privacy budget split equally across three steps, the condition $3200^{-1/6} \le \epsilon$ translates to $\epsilon_{\rm total} \ge 0.4512$. This explains why the empirical coverage reaches the nominal $95\%$ level (within twice the Monte Carlo standard errors) once $\epsilon_{\rm total} \ge 0.6$. 

\section{More on Literature, Background, and Methods}\label{appendix: more background}

Recent research has increasingly focused on uncertainty quantification for private outputs. This includes confidence intervals for ERM estimators \citep{wang2018differentially}, simulation-based inference \citep{awan2025simulation, wang2025optimal}, and private Monte-Carlo based methods \citep{ju2022data, chen2025particle}. Canonical frameworks for private hypothesis testing \citep{awan2023canonical, awan2018differentially, awan2025differentially} and studies on DP-SGD \citep{xie2025online, shi2025non} address inference, efficiency, and adaptivity in practice.

Using the tradeoff functions defined in Section~\ref{sec: privacy background}, we calibrate the pure-DP baselines to the GDP budget used in the simulations.

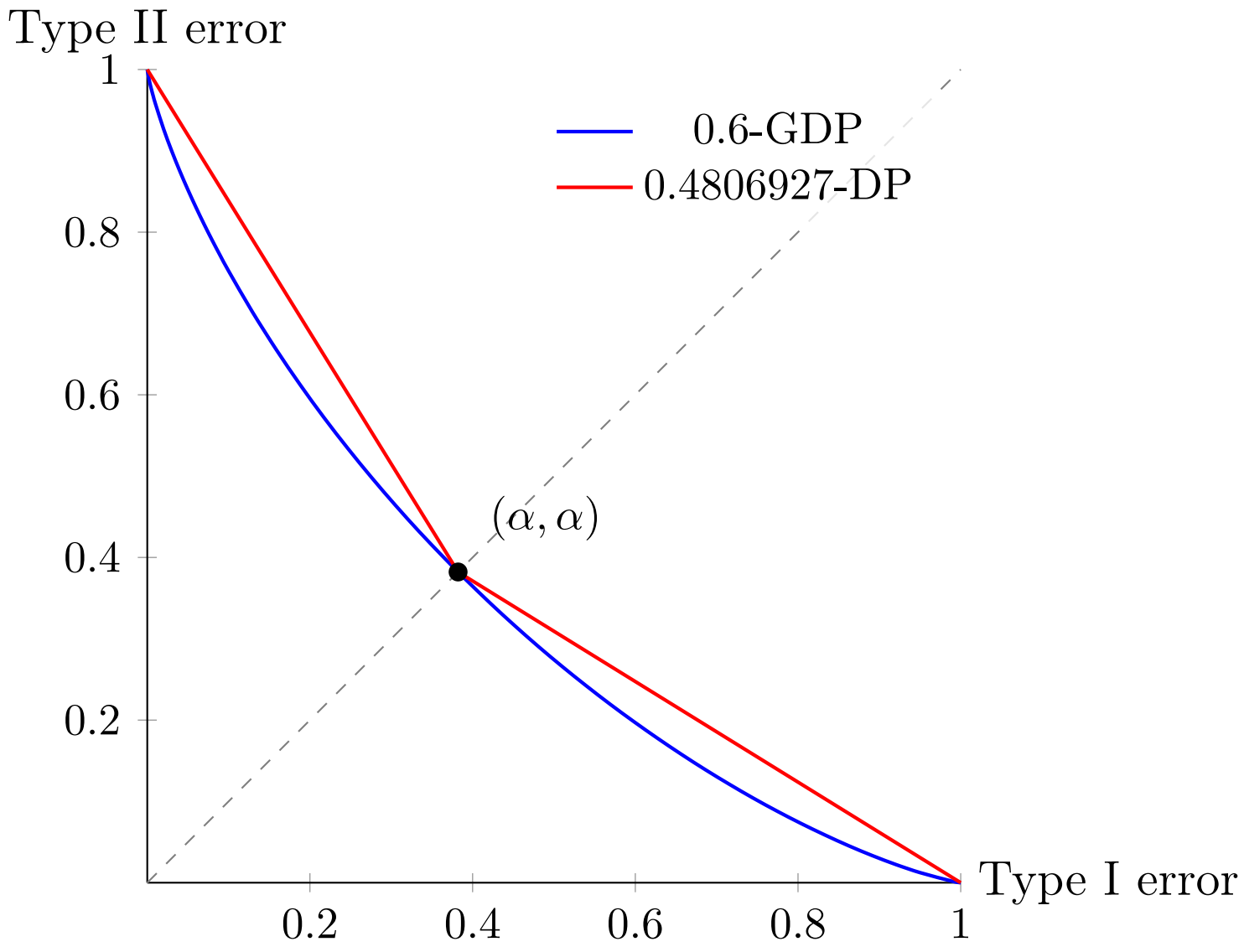
\begin{figure}
\centering
\begin{tikzpicture}
\begin{axis}[
  width=8cm, height=8cm,
  xmin=0, xmax=1, ymin=0, ymax=1,
  xlabel={Type I error},
  ylabel={Type II error},
  axis lines=middle,
  axis line style={-},
  every axis x label/.style={at={(ticklabel* cs:1)}, anchor=west},
  every axis y label/.style={at={(ticklabel* cs:1)}, anchor=south},
  xtick={0,0.2,0.4,0.6,0.8,1},
  ytick={0,0.2,0.4,0.6,0.8,1},
  tick label style={font=\small},
  legend style={at={(0.97,0.97)}, anchor=north east, font=\small, draw=none, fill=white, fill opacity=0.8, text opacity=1},
  clip=false,
]
 
\addplot[blue, thick, smooth] coordinates {
  (0.0000, 1.0000)
  (0.0025, 0.9863)
  (0.0050, 0.9759)
  (0.0075, 0.9666)
  (0.0100, 0.9579)
  (0.0200, 0.9270)
  (0.0300, 0.8999)
  (0.0400, 0.8751)
  (0.0500, 0.8520)
  (0.0600, 0.8302)
  (0.0700, 0.8094)
  (0.0800, 0.7896)
  (0.0900, 0.7706)
  (0.1000, 0.7522)
  (0.1293, 0.7018)
  (0.1586, 0.6555)
  (0.1879, 0.6124)
  (0.2172, 0.5720)
  (0.2466, 0.5340)
  (0.2759, 0.4981)
  (0.3052, 0.4640)
  (0.3345, 0.4315)
  (0.3638, 0.4007)
  (0.3821, 0.3821)
  (0.3931, 0.3712)
  (0.4224, 0.3430)
  (0.4517, 0.3161)
  (0.4810, 0.2903)
  (0.5103, 0.2657)
  (0.5397, 0.2421)
  (0.5690, 0.2195)
  (0.5983, 0.1980)
  (0.6276, 0.1774)
  (0.6569, 0.1577)
  (0.6862, 0.1389)
  (0.7155, 0.1211)
  (0.7448, 0.1041)
  (0.7741, 0.0881)
  (0.8034, 0.0730)
  (0.8328, 0.0588)
  (0.8621, 0.0455)
  (0.8914, 0.0333)
  (0.9207, 0.0222)
  (0.9500, 0.0124)
  (0.9900, 0.0017)
  (1.0000, 0.0000)
};
\addlegendentry{$0.6$-GDP}
 
\addplot[red, thick] coordinates {
  (0.0000, 1.0000)
  (0.3821, 0.3821)
};
\addplot[red, thick, forget plot] coordinates {
  (0.3821, 0.3821)
  (1.0000, 0.0000)
};
\addlegendentry{$0.4806927$-DP}

\addplot[gray, dashed, thin, domain=0:1, samples=2] {x};
 
\node[circle, fill=black, inner sep=1.5pt] at (axis cs:0.3821, 0.3821) {};
\node[anchor=south west, font=\small] at (axis cs:0.40, 0.40) {$(\alpha,\alpha)$};
 
\end{axis}
\end{tikzpicture}
\caption{$0.4806927$-DP implies $0.6$-GDP}
\label{fig:privacy_calibration}
\end{figure}

Based on \citet{awan2024optimizing}, we calculate the tightest pure-DP budget that satisfies $0.6$-GDP. The solution of $G_\epsilon(\alpha)=\alpha$ is $\Phi(-\epsilon/2)$. Now, according to Lemma~2 in \citet{awan2024optimizing}, the DP privacy budget that satisfies $0.6$-GDP is
\begin{align*}
    \log \left( \frac{1-\Phi(\frac{-0.6}{2})}{\Phi(\frac{-0.6}{2})} \right) = 0.4806927.
\end{align*}
Figure~\ref{fig:privacy_calibration} illustrates the calibration target.

\subsection{M-estimation} 

\begin{example}[Consistency of the observed information matrix] 
Let the observed information matrix be 
$$
\hat{I}_{D_n}(\theta)
= \frac{1}{n} \sum_{i=1}^n 
\nabla_\theta \log f_\theta(s_i)\,
\nabla_\theta \log f_\theta(s_i)^{\top}.
$$ 
Assume the following regularity conditions hold for all $\theta$ in a neighborhood of $\theta_0$:  
\begin{enumerate}
    \item $\ex_\theta[\nabla_\theta \log f_\theta(s)] = 0$ and $\ex_\theta[\nabla_\theta \log f_\theta(s)\nabla_\theta \log f_\theta(s)^{\top}] < \infty$;
    \item  $\nabla_\theta \ex_\theta[T(s)] = \ex_\theta[\nabla_\theta T(s)]$ for integrable $T$;
    \item The MLE $\hat{\theta}_n$ is consistent: $\hat{\theta}_n \xrightarrow{p} \theta_0$;
    \item The Fisher information $\fisher(\theta_0) = \ex_{\theta_0}[\nabla_\theta \log f_{\theta_0}(s)\nabla_\theta \log f_{\theta_0}(s)^{\top}]$ exists and is finite.
\end{enumerate}
Then
$$
\hat{I}_{D_n}(\hat{\theta}_n) \xrightarrow{a.s.} I(\theta_0).
$$
\end{example}







\subsection{Competing Methods}\label{appendix: method details}

We describe each competing method in detail, including calibration formulas used in the simulations of Section~\ref{sec: simulation}.

The non-private ERM in \eqref{eq: empirical loss}, with $\Theta=\Theta_B$, serves as the oracle baseline against which the private methods are compared.




The objective perturbation mechanism (Algorithm~1 of \citealt{wang2018differentially}) solves
\begin{align*}
  \hat\theta_{\rm obj} \;=\; \arg\min_{\theta \in \Theta_B} \left\{ F_n(\theta) + \frac{c}{2}\|\theta\|_2^2 + \frac{1}{n} b^\top \theta \right\},
\end{align*}
where $b \sim \mathrm{SpherLap}(\varepsilon'/(2\,\mathrm{grad\_bound}))$ is drawn from the spherical Laplace distribution with rate parameter $\varepsilon'/(2\,\mathrm{grad\_bound})$, and $\varepsilon' = \varepsilon - \log(1 + t/(2nc))$ is the effective privacy budget after accounting for the regularization. Here $t$ is the spectral norm bound on the per-sample Hessian ($t = d/4$ for logistic regression).

We use the canonical regularization sequence $c = c_n = \kappa\,t/(2n(e^\varepsilon - 1))$ with $\kappa = 2$ (Equation~(17) of \citealt{wang2018differentially}), which guarantees $\varepsilon' > 0$ for all $n$ and all $\varepsilon > 0$. If the fixed-$c$ version were used instead (e.g.\ $c = 10^{-3}$), one must verify $c > t/(2n(e^\varepsilon-1))$ for each $n$; failing this condition makes $\varepsilon' \le 0$, forcing a fallback that collapses the mechanism to the non-private ERM.



The noisy gradient descent method of \citet{avella2023differentially} runs $K = 100$ full-batch gradient descent steps with Gaussian perturbations:
\begin{align*}
  \theta_{t+1} \;=\; \mathrm{proj}_{\Theta_B}\!\left(\theta_t - \eta\,\nabla F_n(\theta_t) + \sigma_1\,\xi_t\right), \quad \xi_t \overset{\rm iid}{\sim} \mathcal{N}(0, I_d),
\end{align*}
with step size $\eta = 1$ and noise standard deviation
\begin{align*}
  \sigma_1 \;=\; \frac{2\eta\,C\sqrt{K}}{\epsilon_{\rm main}\,n},
\end{align*}
where $C = B_{\rm grad} = 1$ is the gradient clipping threshold and $\epsilon_{\rm main} = \epsilon_{\rm total}/\sqrt{3}$ is the allocated GDP budget. The method starts from $\theta_0 = \mathbf{0}$ (projected onto $\Theta_B$).

The private sandwich covariance is estimated via two independent Gaussian mechanism releases, each consuming budget $\epsilon_{\rm main}$: one for the empirical Hessian $M_n = n^{-1}\sum_i \nabla^2 f(\theta_K; x_i, y_i)$ with sensitivity $2B_{\rm bar}/n$ ($B_{\rm bar} = 0.25$), and one for the score covariance $Q_n = n^{-1}\sum_i \nabla f(\theta_K; x_i, y_i)\nabla f(\theta_K; x_i, y_i)^\top$ with sensitivity $2C^2/n$. The projected-PSD noise matrices are added to $M_n$ and $Q_n$ and floored at $10^{-6}$ on the eigenvalues. Two variance estimates are reported: Avella-B uses the basic sandwich $\tilde V_n = \tilde M_n^{-1} \tilde Q_n \tilde M_n^{-1}$, and Avella-C adds the GD noise correction
\begin{align*}
  \hat V_n^{\rm corr} \;=\; \tilde V_n + \frac{8\eta^2 C^2 K}{n\,\epsilon_{\rm main}^2}\,I_d,
\end{align*}
which accounts for the variance introduced by the noisy iterates.

\citet{xia2025inference} construct a confidence interval from DP-SGD. We implement their method with $T = n^2$ steps, mini-batch $m = \lfloor\sqrt{n}\rfloor$, and noise $\sigma_1 = C\sqrt{T}/(m\,\epsilon_{\rm main})$. The Polyak--Ruppert average $\bar\theta_T$ serves as the point estimate. The private sandwich covariance is
\begin{align*}
  \hat V_n^{\rm Xia} \;=\; \tilde A_n^{-1} \tilde S_n \tilde A_n^{-1} + \frac{\sigma_1^2}{k_{\rm eff}}\,(\tilde A_n^{-1})^2,
\end{align*}
where $\tilde A_n$ and $\tilde S_n$ are Gaussian mechanism releases of the empirical Hessian and score covariance at $\bar\theta_T$, and $k_{\rm eff} = T/n$ is the effective number of data passes. The CI is the ellipsoidal region $\mathcal{C}_n(\hat V_n^{\rm Xia})$, constructed identically to our method. 

Wang's confidence interval (Algorithm~3 of \citealt{wang2018differentially}) uses three independent $\varepsilon/3$-DP components to privately estimate the point estimate $\tilde\theta$, Hessian $\tilde H$, and score covariance $\tilde\Sigma$. The method draws $m_{\rm MC} = 2000$ Monte Carlo samples
\begin{align*}
  z_i \;=\; \tilde\theta + \frac{1}{\sqrt{n}}\tilde H^{-1}\!\left(G_i + \frac{\beta_i}{\sqrt{n}}\right), \quad G_i \overset{\rm iid}{\sim} \mathcal{N}(0, \tilde\Sigma), \quad \beta_i \overset{\rm iid}{\sim} \mathrm{SpherLap}(\varepsilon'/2),
\end{align*}
and reports per-coordinate quantile intervals. Since this provides only coordinate-wise confidence intervals, we construct a joint confidence region by forming the empirical Mahalanobis ellipsoid from the draws, calibrated to the $95\%$ quantile of the squared Mahalanobis distances. This avoids the conservatism of Bonferroni correction and requires no distributional assumption. The generalised diameter is $2\sqrt{q}\,(\det S)^{1/(2d)}$ where $S$ is the sample covariance of the draws, directly comparable to our ellipsoidal diameter. Wang's CI targets the regularized optimum $\theta_0^\# = \arg\min_\theta\{F(\theta) + (c/2)\|\theta\|_2^2\}$ rather than $\theta_0$, inducing a bias of order $O(c_n) = O(1/n)$.

\subsection{Algorithms} \label{appendix: algorithms}

\begin{algorithm}[t]
\caption{Random-Walk Metropolis--Hastings, $\mathsf{RWMH}(Q, k, \Theta_B, T, T_0)$}\label{alg: M-H sampling for ExpMech}
\begin{algorithmic}[1]
\REQUIRE Objective $Q(\theta)$, concentration parameter $k > 0$, parameter space $\Theta_B$, iterations $T$, burn-in $T_0$
\STATE (Optional) Set proposal scale $\hat{\sigma}_{\rm prop} > 0$; default $\hat{\sigma}_{\rm prop} = \bigl(k\,\lambda_{\min}\bigl(\nabla^2 Q(\hat\theta_n)\bigr)\,d\bigr)^{-1/2}$
\STATE Initialize $\theta^{(0)} \leftarrow 0_d$
\FOR{$t = 1, \ldots, T$}
    \STATE Propose $\theta' = \theta^{(t-1)} + \epsilon$, $\epsilon \sim \mathcal{N}(0, \hat{\sigma}_{\rm prop}^2 I_d)$
    \STATE Set $\alpha = \min\bigl(1, \exp[-k(Q(\theta') - Q(\theta^{(t-1)}))]\bigr)$ if $\theta' \in \Theta_B$, else $\alpha = 0$
    \STATE Draw $u \sim \mathrm{Unif}(0,1)$; set $\theta^{(t)} = \theta'$ if $u \le \alpha$, otherwise $\theta^{(t)} = \theta^{(t-1)}$
\ENDFOR
\ENSURE $\theta^{(T)}$ after discarding the first $T_0$ burn-in samples
\end{algorithmic}
\end{algorithm}

Algorithm~\ref{alg: M-H sampling for ExpMech} summarizes the Metropolis--Hastings procedure used to sample from the exponential mechanism density. The chain is initialized at the origin $0_d$, and the proposal standard deviation is calibrated to the local curvature of the objective at $\hat\theta_n$, scaled by the privacy parameter $k$ and the dimension $d$, to ensure efficient mixing. At each iteration, a Gaussian random walk proposal is accepted or rejected according to the standard Metropolis--Hastings ratio, and the final private estimate $\theta_n^{\rm DP}$ is taken as the last sample after discarding an initial burn-in period of $T_0$ steps.

\begin{algorithm}[t]
\caption{Private Covariance Release \citep{avella2023differentially}, $\mathsf{PrivCov}(f, \theta, D, \epsilon_H, \epsilon_\fisher)$}\label{alg: private covariance} 
\begin{algorithmic}[1]
\REQUIRE Loss $f(\theta; x, y)$, point $\theta \in \Theta_B$, dataset $D = \{(x_i, y_i)\}_{i=1}^n$, budgets $\epsilon_H, \epsilon_\fisher > 0$, floor $\eta > 0$, bounds $t \ge \|\nabla^2 f(\theta; x, y)\|_{\rm op}$ and $R \ge \|\nabla f(\theta; x, y)\|_2$ for all $(x, y) \in \mathcal{X} \times \mathcal{Y}$
\STATE (Optional) Clamp each $(x_i, y_i)$ to $\mathcal{X} \times \mathcal{Y}$ if not already in the domain
\STATE Compute $H_n = n^{-1}\sum_i \nabla^2 f(\theta; x_i, y_i)$ and $\fisher_n = n^{-1}\sum_i \nabla f(\theta; x_i, y_i)\nabla f(\theta; x_i, y_i)^\top$
\STATE Set sensitivities $\Delta_H = 2t/n$ and $\Delta_\fisher = 2R^2/n$
\STATE Let $\mathrm{PSD}_\eta(M) = \sum_j \max(\lambda_j, \eta)\, v_j v_j^\top$, where $\lambda_j, v_j$ are eigenvalues and eigenvectors of matrix $M$
\STATE Draw $G_H, G_\fisher \in \mathbb{R}^{d \times d}$ with i.i.d.\ $\mathcal{N}(0,1)$ entries; set $W_H = (G_H + G_H^\top)/2$ and $W_\fisher = (G_\fisher + G_\fisher^\top)/2$
\STATE Set $\hat{H} = \mathrm{PSD}_\eta\!\left(H_n + \tfrac{\Delta_H}{\epsilon_H} W_H\right)$ and $\hat{\fisher} = \mathrm{PSD}_0\!\left(\fisher_n + \tfrac{\Delta_\fisher}{\epsilon_\fisher} W_\fisher\right)$
\STATE Set $\hat\Sigma_1 = \hat{H}^{-1}\hat{\fisher}\,\hat{H}^{-1}$ and $\hat\Sigma_2 = \hat{H}^{-1}$
\ENSURE Joint release $(\hat\Sigma_1, \hat\Sigma_2)$ satisfying $\sqrt{\epsilon_H^2 + \epsilon_\fisher^2}$-GDP
\end{algorithmic}
\end{algorithm}

Algorithm~\ref{alg: private covariance} describes the procedure for privately releasing the sandwich covariance components $\hat\Sigma_1$ and $\hat\Sigma_2$ at a fixed point $\theta$. The non-private empirical Hessian $H_n$ and score covariance $S_n$ are first computed, and their $\ell_2$-sensitivities are derived from the per-sample gradient and Hessian operator norm bounds. Gaussian noise is added symmetrically to each matrix to satisfy the allocated privacy budgets $\epsilon_H$ and $\epsilon_S$, and the resulting matrices are projected onto the positive semidefinite cone, with $\hat H$ additionally floored at $\eta$ to ensure invertibility. The private estimates $\hat\Sigma_1$ and $\hat\Sigma_2$ are then formed from $\hat H$ and $\hat \fisher$ via the usual sandwich formula. We include a concise proof of its GDP guarantee:
\begin{proof}[Proof of the GDP Guarantee for Algorithm~\ref{alg: private covariance}]
Each matrix $H_n$ and $\fisher_n$ has $\ell_2$ sensitivity $\Delta_H$ and $\Delta_\fisher$ respectively, so adding Gaussian noise with scale $\Delta_H/\epsilon_H$ and $\Delta_\fisher/\epsilon_\fisher$ yields $\epsilon_H$-GDP and $\epsilon_\fisher$-GDP by the Gaussian mechanism \citep{dong2022gaussian}. The projection $\mathrm{PSD}_\eta$ is deterministic post-processing and preserves privacy. Since the two noise draws are independent, the overall guarantee is $\sqrt{\epsilon_H^2 + \epsilon_\fisher^2}$-GDP by composition.
\end{proof}

\section{Joint Asymptotics and Confidence Regions under General Regularization}\label{app: general inference}

This appendix retains all regularization regimes and the general centering conditions. The main-text results are their strongly convex and balanced-regime specializations.

\begin{restatable}[Joint Asymptotic Normality with Exponential Mechanisms: General Regularization]{theorem}{restateThmExpMechJointGeneral}
\label{thm: exp mech joint general}
Under the sampling, convexity, and integrability conditions of Section~\ref{sec: joint exp mech}, together with (A1), (A2$'$), (A3$'$), and (A4$'$), use $\Sigma_1$ as defined in Section~\ref{sec: m estimation setup}. For the strongly convex loss for ExpMech, $\epsilon_n n\to\infty$ gives
\[
\begin{pmatrix}\sqrt n(\hat\theta_n-\theta_0)\\
\epsilon_n n(\theta_n^{DP}-\hat\theta_n)\end{pmatrix} \xrightarrow{d}
N_{2d}\!\left(0,\begin{pmatrix}\Sigma_1&0\\0&(L^2/\mu)H^{-1}\end{pmatrix}\right).
\]
For RegExpMech assume $0<\mu_n=o(n)$ and $c_n=\epsilon_n\sqrt{n\mu_n}\to\infty$, with $\kappa_n$ as in \eqref{eq: joint tuning scales}.
\begin{enumerate}
\item[(i)] If $\kappa_n\to\kappa\in[0,\infty)$, then
\[
\begin{pmatrix}\sqrt n(\hat\theta_n-\theta_0)\\
c_n(\theta_n^{DP}-\hat\theta_n)\end{pmatrix} \xrightarrow{d} N_{2d}\!\left(
\begin{pmatrix}0\\-\kappa H^{-1}\theta_0\end{pmatrix},
\begin{pmatrix}\Sigma_1&0\\0&L^2H^{-1}\end{pmatrix}\right).
\]
Recall that the cases $\kappa=0$ and $0<\kappa<\infty$ are the mechanism-variance-dominated and balanced regimes.
\item[(ii)] If $\kappa_n\to\infty$, then
\[
\begin{pmatrix}\sqrt n(\hat\theta_n-\theta_0)\\
(n/\mu_n)(\theta_n^{DP}-\hat\theta_n)\end{pmatrix}
\xrightarrow{d}\begin{pmatrix}X\\-H^{-1}\theta_0\end{pmatrix},
\qquad X\sim N_d(0,\Sigma_1).
\]
\item[(iii)] If \eqref{eq: conditional correction tuning} holds,
then with $A_n=I_d+(\mu_n/n)H^{-1}$,
\[
\begin{pmatrix}\sqrt n(\hat\theta_n-\theta_0)\\
c_n(A_n\theta_n^{DP}-\hat\theta_n)\end{pmatrix} \xrightarrow{d} N_{2d}\!\left(0,
\begin{pmatrix}\Sigma_1&0\\0&L^2H^{-1}\end{pmatrix}\right).
\]
\end{enumerate}
\end{restatable}

Part (iii) follows by combining the conditional bias-correction result with the data CLT through Theorem~\ref{thm: joint convergence}. Its conclusion also holds under the weaker sufficient conditions
\begin{equation}\label{eq: stochastic correction tuning}
\epsilon_n\mu_n^{3/2}=o(n),\qquad \epsilon_n\mu_n^{5/2}=o(n^{3/2}).
\end{equation}
The stochastic expansion in the proof of Theorem~\ref{thm: confidence region general} establishes this extension by controlling the remaining bias in probability.

The regimes above compare regularization bias with mechanism noise. To recover the non-private limiting distribution, both errors must also be negligible relative to the sampling error, whose scale is $n^{-1/2}$. For the uncorrected RegExpMech estimator, sufficient conditions are $\epsilon_n\sqrt{\mu_n}\to\infty$ and $\mu_n=o(n^{1/2})$. At the balancing choice $\mu_n=(n/\epsilon_n^2)^{1/3}$, both additional error scales are $(\epsilon_n n)^{-2/3}$, so these conditions reduce to $\epsilon_n n^{1/4}\to\infty$. Thus, balancing regularization bias and mechanism noise alone does not necessarily recover the non-private limiting distribution. For the strongly convex mechanism, the corresponding condition is $\epsilon_n\sqrt n\to\infty$. 

For fixed $\epsilon_n=\epsilon>0$, removing $\epsilon$ from the mechanism normalizations gives limiting covariance $L^2H^{-1}/\epsilon^2$ in the convex case and $L^2H^{-1}/(\mu\epsilon^2)$ in the strongly convex case. The variance-dominated regime corresponds to $\mu_n=o(n^{1/3})$, the balanced regime to $\mu_n/n^{1/3}\to c$ for some $c>0$, and the bias-dominated regime to $\mu_n/n^{1/3}\to\infty$. In particular, choosing $\mu_n=n^{1/3}$ gives limiting mean $-H^{-1}\theta_0$ under the normalization $n^{2/3}$. Since $\zeta<1/10$, the conditional correction conditions hold whenever $\mu_n=o(n^{3/5})$, recovering the fixed-budget bias-correction result.

For RegExpMech, write
\begin{equation}\label{eq: sampling comparison scales}
s_n=\sqrt{\frac1n+\frac1{\epsilon_n^2n\mu_n}}.
\end{equation}
The centering conditions compare the remaining bias with this combined error scale.
\begin{restatable}[Asymptotic Confidence Region: General Regularization]{theorem}{restateThmConfidenceRegionGeneral}\label{thm: confidence region general}
Under the sampling, convexity, and integrability conditions of Section~\ref{sec: joint exp mech}, together with (A1), (A2$'$), (A3$'$), and (A4$'$), assume $H^{-1}\mathcal I H^{-1}\succ0$. Let $\hat\Sigma_{1,n}$ and $\hat\Sigma_{2,n}$ be private symmetric consistent estimators of $H^{-1}\mathcal I H^{-1}$ and $H^{-1}$, respectively, and let $\epsilon_n>0$ be a deterministic sequence.
\begin{itemize}
\item For the strongly convex loss on ExpMech in Theorem~\ref{thm: strong conv utility}, suppose $\epsilon_n n\to\infty$ and set
\[
\hat V_n=\hat\Sigma_{1,n}+\frac{L^2}{\mu\epsilon_n^2n}\hat\Sigma_{2,n},\qquad \tilde\theta_n=\theta_n^{DP}.
\]
\item For RegExpMech in Lemma~\ref{lem: regularized convex utility}, suppose $0<\mu_n=o(n)$ and $\epsilon_n\sqrt{n\mu_n}\to\infty$. Set
\[
\hat V_n=\hat\Sigma_{1,n}+\frac{L^2}{\epsilon_n^2\mu_n}\hat\Sigma_{2,n},\qquad \hat A_n=I_d+\frac{\mu_n}{n}\hat\Sigma_{2,n},
\]
\begin{equation}\label{eq: private correction rate}
\tilde\theta_n=
\begin{cases}
\theta_n^{DP}, & \dfrac{\mu_n}{ns_n}\to0,\\[2ex]
\hat A_n\theta_n^{DP}, & \dfrac{\mu_n^2}{n^2s_n}\to0 \quad\text{and}\quad \dfrac{\mu_n}{s_n}\|\hat\Sigma_{2,n}-H^{-1}\|=o_p(n).
\end{cases}
\end{equation}
\end{itemize}
Then $\sqrt n\,\hat V_n^{-1/2}(\tilde\theta_n-\theta_0)\dlim N_d(0,I_d)$, and $\mathcal C_n(\hat V_n)=\left\{\theta\in\mathbb R^d:n(\tilde\theta_n-\theta)^\top\hat V_n^{-1}(\tilde\theta_n-\theta)\le\chi^2_{d,1-\alpha}\right\}$ is an asymptotic $(1-\alpha)$ confidence region for $\theta_0$.
\end{restatable}

Using $\theta_n^{DP}$ as the center of the confidence region gives asymptotic coverage $1-\alpha$ when $\mu_n/(ns_n)\to0$, provided the other assumptions of Theorem~\ref{thm: confidence region general} hold. This condition makes the regularization bias negligible relative to the combined sampling and mechanism error scale $s_n$. Bias correction permits a larger regularization parameter, provided the remaining bias and the error from estimating the correction are both negligible. If both choices satisfy their conditions, either center may be used.

For fixed $\epsilon_n=\epsilon>0$, the uncorrected choice includes $\mu_n=o(n^{1/3})$, subject to mechanism concentration. The corrected choice includes $\mu_n=n^\gamma$ with $1/3\le\gamma<3/5$, provided \eqref{eq: private correction rate} holds. In this range, that condition is equivalent to $\mu_n\|\hat\Sigma_{2,n}-H^{-1}\|=o_p(n^{1/2})$. Consistency suffices for $\gamma\le1/2$; larger values require a convergence rate for the estimated correction.

\section{Proofs}

Throughout the proofs, $\epsilon_n>0$ and $\mu_n>0$ are the deterministic sequences specified in each result. We write $\Sigma_2=H^{-1}$ and, in the random-data analysis, $\Sigma_1=H^{-1}\mathcal I H^{-1}$. The single matrix $H^{-1}$ suffices in the conditional analysis, where only the mechanism-noise covariance appears; the full sandwich $\Sigma_1$ enters once data randomness is reintroduced. These two matrices coincide under the information equality $\mathcal I=H$.

Fix a dataset sequence satisfying (A1)--(A4). We first establish a fixed neighborhood containing the relevant minimizers on which the empirical Hessians have a common positive lower bound. Write $\overline M_n=n^{-1}\sum_{i=1}^nM(s_i)$ and $C=\sup_n\overline M_n<\infty$, where finiteness follows from (A2). Since $\mathcal U$ is an open neighborhood of $\theta_0$ contained in $\operatorname{int}(\Theta)$, we can choose $r_0>0$ such that $K=\overline B(\theta_0,r_0)\subset\mathcal U$ and $Cr_0<\lambda_{\min}(H)/4$. The common bound $C$ allows this radius to be fixed independently of $n$. By (A3), $\|\nabla^2F_n(\theta_0)-H\|<\lambda_{\min}(H)/4$ for all sufficiently large $n$. Thus, \eqref{eq: local Hessian control} gives
\[
\sup_{\theta\in K}\|\nabla^2F_n(\theta)-H\|
\le \|\nabla^2F_n(\theta_0)-H\|+Cr_0
<\frac{\lambda_{\min}(H)}{2}.
\]
Consequently, with $m=\lambda_{\min}(H)/2>0$, we have $\nabla^2F_n(\theta)\succeq mI_d$ throughout $K$ for all sufficiently large $n$. In the regularized case,
\[
\nabla^2F_n^\#(\theta)
=\nabla^2F_n(\theta)+\frac{\mu_n}{n}I_d
\succeq mI_d
\]
on the same ball. By (A4), the relevant minimizers eventually lie in the interior of $K$, and their joining segment remains in $K$ by convexity. These facts justify the first-order conditions, Taylor expansions, and bounds on inverse Hessians used below. The closed ball is compact, although compactness is not needed for the Hessian lower bound. We require uniformly positive curvature on this fixed ball, rather than uniform convergence of its Hessian to the constant matrix $H$.

\restatePropDpGuaranteeOfExpMech*

With the convexity and sensitivity conditions imposed over $\mathcal S^n$, Proposition~\ref{prop: dp guarantee of exp mech} follows from the result below.
\begin{restatable}[Theorem 13, \citet{gopi2022private}]{lemma}{restateLemmaGopiDpGuarantee}\label{lemma: gopi dp guarantee}
Let $\mathcal K\subseteq\mathbb R^d$ be convex, and let $F,\widetilde F$ be $\mu$-strongly convex on $\mathcal K$. Let $P,Q$ have densities proportional to $e^{-F}$ and $e^{-\widetilde F}$, respectively. If $F-\widetilde F$ is $L$-Lipschitz on $\mathcal K$, then, for all $z\in[0,1]$,
\[
T(P\Vert Q)(z)\ge T\left(N(0,1)\Vert N\left(\frac{L}{\sqrt\mu},1\right)\right)(z).
\]
\end{restatable}

\begin{proof}
The privacy guarantee follows from the GDP properties of log-concave distributions in Lemma~\ref{lemma: gopi dp guarantee}. We apply it to neighboring datasets $D,D'$ to derive the scale parameter $k_n$ for a target $\epsilon_n$-GDP guarantee. Convexity and the stated strong-convexity constants in this privacy argument must hold for every neighboring dataset in $\mathcal S^n$, and the sensitivity constant $L$ is common to the data domain.

\emph{Strongly Convex Loss.}
The exponent $k_nF_n(\theta;D)$ is $k_n\mu$-strongly convex, while $k_nF_n(\theta;D)-k_nF_n(\theta;D')$ is $k_nL/n$-Lipschitz. Lemma~\ref{lemma: gopi dp guarantee} therefore gives GDP parameter
\[
\frac{k_nL/n}{\sqrt{k_n\mu}}=\frac{\sqrt{k_n}L}{n\sqrt\mu}.
\]
Equating this expression to $\epsilon_n$ gives $k_n=\epsilon_n^2\mu n^2/L^2$.

\emph{Convex Loss with Regularization.}
The regularizer makes $F_n^\#(\theta;D)=F_n(\theta;D)+(\mu_n/2n)\|\theta\|^2$ $\mu_n/n$-strongly convex. It is identical for $D,D'$, so it cancels in their difference, which remains $L/n$-Lipschitz. The GDP parameter is consequently
\[
\frac{k_nL/n}{\sqrt{k_n\mu_n/n}}=\frac{\sqrt{k_n}L}{\sqrt{n\mu_n}}.
\]
Setting it equal to $\epsilon_n$ gives $k_n=\epsilon_n^2\mu_n n/L^2$. This proves the calibration in both cases.
\end{proof}

The following example calculates the Lipschitz constants used in Section~\ref{sec: simulation}.
\begin{example}[Lipschitz constant calculation]\label{eg: logistic_LM_Lipshitz}
Let $s=(x,y)$ and use the Euclidean norm. Suppose the public feature bound is $\|x\|\le R$.
\begin{itemize}
\item For logistic loss, $y\in\{0,1\}$ and $\nabla f_\theta(x,y)=(\sigma(x^\top\theta)-y)x$. Hence $\|\nabla f_\theta(x,y)\|\le R$ and
\[
\|\nabla f_\theta(s)-\nabla f_\theta(s')\|\le2R.
\]
Thus $L=2R$ is valid. Also $\nabla^2f_\theta(x,y)=\sigma(x^\top\theta)(1-\sigma(x^\top\theta))xx^\top$ has norm at most $R^2/4$.
\item For squared-error loss, suppose additionally $|y|\le Y$ and $\|\theta\|\le B$. Then $\nabla f_\theta(x,y)=(x^\top\theta-y)x$ gives $\|\nabla f_\theta(x,y)\|\le(RB+Y)R$, so $L=2(RB+Y)R$ is valid. Its Hessian $xx^\top$ has norm at most $R^2$.
\end{itemize}
These gradient bounds imply Lipschitz loss differences by integration along line segments in the convex parameter space. Taking $R=\sqrt d$ and $Y=1$ recovers the constants used in the simulations. The squared-error bound requires bounded responses and parameters as well as bounded features; the logistic bound only requires bounded features.
\end{example}

\restateThmStrongConvUtility*
\begin{proof}
Set $r_n=\epsilon_n n\to\infty$ and $Z_n=r_n(\theta_n^{DP}-\hat\theta_n)$. Under the change of variables $\theta=\hat\theta_n+z/r_n$, the Jacobian and the factor $\exp\{-k_nF_n(\hat\theta_n)\}$ cancel upon normalization. Thus $Z_n$ has density $\pi_n(z)=q_n(z)/C_n$, where
\[
q_n(z)=\mathbf1\{\hat\theta_n+z/r_n\in\Theta\}
\exp\left\{-\frac{\mu r_n^2}{L^2}
\left[F_n(\hat\theta_n+z/r_n)-F_n(\hat\theta_n)\right]\right\},
\qquad C_n=\int_{\mathbb R^d}q_n(z)\,dz,
\]
with $q_n$ defined to be zero outside its support.

By (A4) and the interiority of $\theta_0$, $\hat\theta_n$ is eventually interior, so $\nabla F_n(\hat\theta_n)=0$. For $z$ in any fixed compact set, the support indicator is eventually one and the segment $\hat\theta_n+tz/r_n$, $t\in[0,1]$, lies in $\mathcal U$. Integral Taylor expansion gives
\[
r_n^2\left[F_n(\hat\theta_n+z/r_n)-F_n(\hat\theta_n)\right]
=\int_0^1(1-t)z^\top
\nabla^2F_n(\hat\theta_n+tz/r_n)z\,dt.
\]
Writing $C=\sup_n\overline M_n<\infty$, the local Hessian control yields, for every fixed $R<\infty$,
\[
\sup_{\substack{\|z\|\le R\\t\in[0,1]}}
\|\nabla^2F_n(\hat\theta_n+tz/r_n)-H\|
\le \|\nabla^2F_n(\theta_0)-H\|
+C\|\hat\theta_n-\theta_0\|+\frac{CR}{r_n}
\longrightarrow0.
\]
Hence the objective gap multiplied by $r_n^2$ converges to $\frac12z^\top Hz$, and continuity of the exponential gives
\[
q_n(z)\longrightarrow q(z)
=\exp\left\{-\frac{\mu}{2L^2}z^\top Hz\right\}.
\]

To control the normalizing constants, global $\mu$-strong convexity and the first-order condition imply, for every feasible $z$,
\[
F_n(\hat\theta_n+z/r_n)-F_n(\hat\theta_n)
\ge \frac{\mu}{2r_n^2}\|z\|^2.
\]
Therefore $q_n(z)\le\exp\{-\mu^2\|z\|^2/(2L^2)\}$ for all sufficiently large $n$, an integrable bound independent of $n$. Dominated convergence gives $C_n\to C_\infty=\int q(z)\,dz\in(0,\infty)$, since $H\succ0$. Consequently, $\pi_n(z)\to q(z)/C_\infty$, the density of $N_d(0,(L^2/\mu)H^{-1})$. Since $\pi_n(z)\to\pi(z)$ pointwise and both are probability densities, Scheff\'e's theorem gives
\[
d_{\mathrm{TV}}(\mathcal L(Z_n),N_d(0,(L^2/\mu)H^{-1}))
=\frac12\int_{\mathbb R^d}|\pi_n(z)-\pi(z)|\,dz\longrightarrow0,
\]
which also implies the convergence in distribution.
\end{proof}

\restateLemRegularizedConvexUtility*
\begin{proof}
We treat regularization bias and mechanism randomness separately.

\emph{Part 1: Regularization Bias.}
Write $\lambda_n=\mu_n/n\to 0$ and $\Delta_n=\theta_n^\#-\hat\theta_n$. By (A4), both minimizers converge to the interior point $\theta_0$, so they are eventually interior and their joining segment lies in $\mathcal U$. Their first-order conditions are therefore $\nabla F_n(\hat\theta_n)=0$ and $\nabla F_n(\theta_n^\#)+\lambda_n\theta_n^\#=0$. Applying the fundamental theorem of calculus along their joining segment gives $\nabla F_n(\theta_n^\#)-\nabla F_n(\hat\theta_n)=J_n\Delta_n$, where $J_n=\int_0^1\nabla^2 F_n(\hat\theta_n+t\Delta_n)\,dt$. Substituting the first-order conditions and $\theta_n^\#=\hat\theta_n+\Delta_n$ yields the exact bias identity
\begin{equation}\label{eq: deterministic bias identity}
\begin{aligned}
(J_n+\lambda_n I_d)\Delta_n
=\nabla F_n(\theta_n^\#)-\nabla F_n(\hat\theta_n)+\lambda_n(\theta_n^\#-\hat\theta_n)
=-\lambda_n\hat\theta_n.
\end{aligned}
\end{equation}

We first establish a bound on $\Delta_n$ using consistency alone. By \eqref{eq: local Hessian control},
\begin{equation}\label{eq: integrated Hessian control}
\begin{aligned}
\|J_n-H\|
&\le \|\nabla^2F_n(\theta_0)-H\|+\int_0^1\|\nabla^2F_n(\hat\theta_n+t\Delta_n)-\nabla^2F_n(\theta_0)\|\,dt\\
&\le \|\nabla^2F_n(\theta_0)-H\|+\overline M_n\bigl(\|\hat\theta_n-\theta_0\|+\|\Delta_n\|\bigr).
\end{aligned}
\end{equation}
The right-hand side tends to zero: (A3) gives convergence of the first term, (A2) gives boundedness of $\overline M_n$, and (A4) gives $\hat\theta_n\to\theta_0$ and $\Delta_n\to 0$. Since $J_n+\lambda_nI_d\to H\succ0$, its inverse converges to $H^{-1}$ and is therefore bounded. Also, $\hat\theta_n\to\theta_0$ implies that $\hat\theta_n$ is bounded. Thus the two boundedness and \eqref{eq: deterministic bias identity} gives
\[
\|\Delta_n\|
= \|\lambda_n(J_n+\lambda_nI_d)^{-1}\hat\theta_n\|
\le \lambda_n\|(J_n+\lambda_nI_d)^{-1}\|\,\|\hat\theta_n\|
=\lambda_n O(1)O(1)
=O(\lambda_n).
\]
Substituting this bound and the rates $o(n^{-1/2+\zeta})$ in (A3)--(A4) into
\eqref{eq: integrated Hessian control} yields
\begin{align*}
\|J_n-H\|
&\le \|\nabla^2F_n(\theta_0)-H\|+\overline M_n\bigl(\|\hat\theta_n-\theta_0\|+\|\Delta_n\|\bigr)\\
&=o(n^{-1/2+\zeta})+O(1)\bigl[o(n^{-1/2+\zeta})+O(\lambda_n)\bigr]\\
&=O(n^{-1/2+\zeta}+\lambda_n).
\end{align*}
Since both inverses are bounded, the matrix inverse identity gives
\[
\begin{aligned}
\|(J_n+\lambda_nI_d)^{-1}-H^{-1}\|
&\le \|(J_n+\lambda_nI_d)^{-1}\|
\bigl(\|J_n-H\|+\lambda_n\bigr)\|H^{-1}\|\\
&=O(1)O(n^{-1/2+\zeta}+\lambda_n)O(1).
\end{aligned}
\]
Together with $\hat\theta_n=\theta_0+O(n^{-1/2+\zeta})$, the bias identity therefore gives
\begin{align*}
\Delta_n
&=-\lambda_n
\bigl[H^{-1}+O(n^{-1/2+\zeta}+\lambda_n)\bigr]
\bigl[\theta_0+O(n^{-1/2+\zeta})\bigr]\\
&=-\lambda_nH^{-1}\theta_0
+O(\lambda_n n^{-1/2+\zeta}+\lambda_n^2)\\
&=-\frac{\mu_n}{n}H^{-1}\theta_0
+O\left(\frac{\mu_n}{n^{3/2-\zeta}}+\frac{\mu_n^2}{n^2}\right).
\end{align*}
Dividing by $\lambda_n=\mu_n/n$ gives
$(n/\mu_n)\Delta_n=-H^{-1}\theta_0
+O(n^{-1/2+\zeta}+\lambda_n)\to-H^{-1}\theta_0$.

\emph{Part 2: Mechanism Randomness.}
Define $r_n=\epsilon_n\sqrt{n\mu_n}\to\infty$ and $Z_n=r_n(\theta_n^{DP}-\theta_n^\#)$. The density of $Z_n$ is $q_n(z)/C_n$, where
\[
\begin{gathered}
q_n(z)=\mathbf1\{\theta_n^\#+z/r_n\in\Theta\}e^{-g_n(z)},\qquad C_n=\int q_n(z)\,dz,\\
g_n(z)=\frac{r_n^2}{L^2}\left[F_n^\#(\theta_n^\#+z/r_n)-F_n^\#(\theta_n^\#)\right].
\end{gathered}
\]
The function $q_n$ is defined to be zero outside its support. Fix a compact set of $z$ and put $h_n=z/r_n$. The indicator is eventually one uniformly on this set. The first-order condition and Taylor expansion give
\[
g_n(z)=\frac1{L^2}\int_0^1(1-t)z^\top\left[\nabla^2F_n(\theta_n^\#+th_n)+\frac{\mu_n}{n}I_d\right]z\,dt\longrightarrow\frac1{2L^2}z^\top Hz,
\]
uniformly on that compact set, by \eqref{eq: local Hessian control}, consistency, and $\mu_n=o(n)$.

To apply dominated convergence, we need an integrable bound independent of $n$. The global strong-convexity constant $\mu_n/n$ tends to zero, so we instead use the fixed positive local curvature established at the start of the appendix. On a fixed ball $K=\overline B(\theta_0,r_0)\subset\mathcal U$, we have $\nabla^2F_n^\#(\theta)\succeq mI_d$ for some $m>0$ and all sufficiently large $n$. Since $\theta_n^\#\to\theta_0$, choosing $r=r_0/2$ ensures that $\overline B(\theta_n^\#,r)\subset K$ eventually.

Here $r$ is a fixed neighborhood radius, whereas $r_n=\epsilon_n\sqrt{n\mu_n}$ rescales the mechanism error. For a parameter point $\theta\in\overline B(\theta_n^\#,r)$ inside the ball, the corresponding rescaled point is $z=r_n(\theta-\theta_n^\#)$. The entire segment $\theta_n^\#+t(\theta-\theta_n^\#)$, $t\in[0,1]$, lies in $K$. Using $\nabla F_n^\#(\theta_n^\#)=0$ and the Hessian lower bound, integral Taylor expansion gives
\[
\begin{aligned}
F_n^\#(\theta)-F_n^\#(\theta_n^\#)
&=\int_0^1(1-t)(\theta-\theta_n^\#)^\top
\nabla^2F_n^\#\bigl(\theta_n^\#+t(\theta-\theta_n^\#)\bigr)
(\theta-\theta_n^\#)\,dt\\
&\ge \int_0^1(1-t)m\|\theta-\theta_n^\#\|^2\,dt
=\frac{m}{2}\|\theta-\theta_n^\#\|^2.
\end{aligned}
\]
For a feasible point $\theta\in\Theta$ outside the ball, write
$d=\|\theta-\theta_n^\#\|>r$ and define
\[
\theta_r=\theta_n^\#+\frac r d(\theta-\theta_n^\#).
\]
This point lies on the segment from $\theta_n^\#$ to $\theta$, at distance $r$ from $\theta_n^\#$, where the quadratic bound applies. Since
\[
\theta_r=\left(1-\frac r d\right)\theta_n^\#+\frac r d\theta,
\]
convexity gives
\[
F_n^\#(\theta_r)-F_n^\#(\theta_n^\#)
\le \frac r d
\left[F_n^\#(\theta)-F_n^\#(\theta_n^\#)\right].
\]
The left-hand side is at least $mr^2/2$ by the quadratic bound. Rearranging therefore yields
\[
F_n^\#(\theta)-F_n^\#(\theta_n^\#)
\ge \frac d r\frac{mr^2}{2}
=\frac{mr}{2}\|\theta-\theta_n^\#\|.
\]
For the parameter point $\theta=\theta_n^\#+z/r_n$ corresponding to $z$, we have
$\|\theta-\theta_n^\#\|=\|z\|/r_n$. Recall that
\[
g_n(z)=\frac{r_n^2}{L^2}
\left[F_n^\#(\theta)-F_n^\#(\theta_n^\#)\right],
\qquad q_n(z)=e^{-g_n(z)}
\]
on the support. If $\|z\|\le rr_n$, this point lies inside the ball, so the quadratic bound gives
\[
g_n(z)\ge
\frac{r_n^2}{L^2}\frac m2\left(\frac{\|z\|}{r_n}\right)^2
=\frac{m}{2L^2}\|z\|^2.
\]
If $\|z\|>rr_n$, the point lies outside the ball, so the linear bound gives
\[
g_n(z)\ge
\frac{r_n^2}{L^2}\frac{mr}{2}\frac{\|z\|}{r_n}
=\frac{mr r_n}{2L^2}\|z\|.
\]
Since $q_n(z)=e^{-g_n(z)}$, these lower bounds on $g_n$ yield
\[
q_n(z)\le
\begin{cases}
\exp\{-m\|z\|^2/(2L^2)\},&\|z\|\le rr_n,\\
\exp\{-mr r_n\|z\|/(2L^2)\},&\|z\|>rr_n.
\end{cases}
\]
The bound also holds outside the support, where $q_n=0$. Since $r_n\ge1$ eventually, we obtain the fixed integrable envelope
\[
q_n(z)\le
\exp\{-m\|z\|^2/(2L^2)\}
+\exp\{-mr\|z\|/(2L^2)\},
\]
which permits dominated convergence. Dominated convergence gives $C_n\to\int\exp\{-z^\top Hz/(2L^2)\}\,dz\in(0,\infty)$. Scheff\'e's theorem now gives convergence in total variation, and therefore
\[
\epsilon_n\sqrt{n\mu_n}(\theta_n^{DP}-\theta_n^\#)\dlim N_d(0,L^2H^{-1}).
\]
\end{proof}

\restateThmRegularizedConvexUtility*
\begin{proof}
Let $r_n=\epsilon_n\sqrt{n\mu_n}$ and define
\[
Z_n=r_n(\theta_n^{DP}-\theta_n^\#),
\qquad
B_n=\frac n{\mu_n}(\theta_n^\#-\hat\theta_n).
\]
Lemma~\ref{lem: regularized convex utility} gives
\[
Z_n\dlim N_d(0,L^2H^{-1}),
\qquad B_n\to-H^{-1}\theta_0,
\]
where $B_n$ is deterministic because the dataset sequence is fixed.
The exact error decomposition is
\[
\theta_n^{DP}-\hat\theta_n
=\frac{Z_n}{r_n}+\frac{\mu_n}{n}B_n.
\]
Thus the mechanism-noise scale is $1/r_n$, while the
first-order bias scale is $\mu_n/n$. Their ratio is
\[
\kappa_n=\frac{\mu_n/n}{1/r_n}
=\frac{\epsilon_n\mu_n^{3/2}}{\sqrt n}.
\]

\emph{Case 1: Variance-Dominated Regime.}
If $\kappa_n\to0$, the bias scale is smaller than the
mechanism-noise scale. Multiplying the error decomposition by $r_n$ gives
\[
r_n(\theta_n^{DP}-\hat\theta_n)
=Z_n+\kappa_nB_n.
\]
Since $B_n$ converges to a finite vector, it is bounded, and hence
$\kappa_nB_n\to0$. Slutsky's theorem therefore gives
\[
\epsilon_n\sqrt{n\mu_n}(\theta_n^{DP}-\hat\theta_n)
\dlim N_d(0,L^2H^{-1}).
\]

\emph{Case 2: Balanced Regime.}
If $\kappa_n\to\kappa\in(0,\infty)$, the two scales are comparable.
Under the same normalization,
\[
r_n(\theta_n^{DP}-\hat\theta_n)
=Z_n+\kappa_nB_n,
\qquad
\kappa_nB_n\to-\kappa H^{-1}\theta_0.
\]
Thus the bias remains as a deterministic shift, while the mechanism
randomness retains its Gaussian limit. Slutsky's theorem gives
\[
\epsilon_n\sqrt{n\mu_n}(\theta_n^{DP}-\hat\theta_n)
\dlim N_d(-\kappa H^{-1}\theta_0,L^2H^{-1}).
\]

\emph{Case 3: Bias-Dominated Regime.}
If $\kappa_n\to\infty$, the bias scale is larger than the
mechanism-noise scale. We therefore divide by the bias scale
$\mu_n/n$, rather than by the noise scale:
\[
\frac n{\mu_n}(\theta_n^{DP}-\hat\theta_n)
=\frac{n}{\mu_n r_n}Z_n+B_n
=\frac{Z_n}{\kappa_n}+B_n.
\]
Convergence in distribution implies $Z_n=O_p(1)$.
Since $1/\kappa_n\to0$, it follows that
$Z_n/\kappa_n\to0$ in probability. Together with
$B_n\to-H^{-1}\theta_0$, this yields
\[
\frac n{\mu_n}(\theta_n^{DP}-\hat\theta_n)
\xrightarrow{P}-H^{-1}\theta_0.
\]
If $\theta_0=0$, the leading bias vanishes and this limit is zero;
the regime names refer to the comparison of the two scales.
\end{proof}

\restateCorConditionalBiasCorrection*
\begin{proof}
Lemma~\ref{lem: regularized convex utility} identifies the leading regularization bias as $-(\mu_n/n)H^{-1}\theta_0$. To cancel it, we would add $(\mu_n/n)H^{-1}\theta_0$. Since $\theta_0$ is unknown, we substitute $\theta_n^{DP}$, giving
\[
\widetilde\theta_n
=\theta_n^{DP}+\frac{\mu_n}{n}H^{-1}\theta_n^{DP}.
\]
The lemma and consistency of $\theta_n^\#$ imply $\theta_n^{DP}\to\theta_0$ in probability, so the added term is $O_p(\mu_n/n)$, matching the leading bias scale. We now verify that the remaining error is negligible at the mechanism-noise scale.

Separating the mechanism noise from the regularization shift gives the exact decomposition
\[
\begin{aligned}
\widetilde\theta_n-\hat\theta_n
=\left(I_d+\frac{\mu_n}{n}H^{-1}\right) (\theta_n^{DP}-\theta_n^\#)+(\theta_n^\#-\hat\theta_n) +\frac{\mu_n}{n}H^{-1}\theta_n^\#.
\end{aligned}
\]
By the lemma and (A4), $\theta_n^\#-\theta_0=(\theta_n^\#-\hat\theta_n)+(\hat\theta_n-\theta_0)=O\left(n^{-1/2+\zeta}+\frac{\mu_n}{n}\right)$. Thus the last two terms satisfy
\[
\begin{aligned}
(\theta_n^\#-\hat\theta_n)+\frac{\mu_n}{n}H^{-1}\theta_n^\#
&=-\frac{\mu_n}{n}H^{-1}\theta_0
+\frac{\mu_n}{n}H^{-1}\theta_0+O\left(\frac{\mu_n}{n^{3/2-\zeta}}
+\frac{\mu_n^2}{n^2}\right)\\
&=O\left(\frac{\mu_n}{n^{3/2-\zeta}}
+\frac{\mu_n^2}{n^2}\right).
\end{aligned}
\]
Multiplying the decomposition by the mechanism normalization therefore gives
\[
\begin{aligned}
\epsilon_n\sqrt{n\mu_n}(\widetilde\theta_n-\hat\theta_n)
=\left(I_d+\frac{\mu_n}{n}H^{-1}\right)\epsilon_n\sqrt{n\mu_n}(\theta_n^{DP}-\theta_n^\#)+O\left(\frac{\epsilon_n\mu_n^{3/2}}{n^{1-\zeta}}+\frac{\epsilon_n\mu_n^{5/2}}{n^{3/2}}
\right).
\end{aligned}
\]
The remainder tends to zero by \eqref{eq: conditional correction tuning}, and the multiplying matrix tends to $I_d$ because $\mu_n=o(n)$. The mechanism limit in the lemma and Slutsky's theorem yield the asserted centered Gaussian limit.
\end{proof}

\restateThmJointConvergence*
\begin{proof}
Motivated by \cite{awan2025one}, our proof hinges on a constructive coupling argument that leverages the Skorokhod Representation Theorem to explicitly separate the two sources of randomness—one from the data generating process and one from the mechanism—using independent uniform random variables. 

Recall that $G = \{ (\mathbf{s}_n)_{n=1}^\infty: g_n(\mathbf{s}_n) \to c \in \mathbb{R}^m\}$ is the collection of ``good'' triangular datasets/arrays with the regularity conditions hold. In the remainder of the proof, we focus on such datasets that have good convergence properties.

We first representing the statistical error and convergence of $g_n$ via $U_1$. By assumption (B1) and Slutsky's theorem, we have
$$
\begin{aligned}
\begin{pmatrix}
    n^a(\hat{\theta}_n(D_n) - \theta_0) \\  g_n(D_n)
\end{pmatrix}    
\dlim 
\begin{pmatrix}
    X \\ c
\end{pmatrix}
\end{aligned}
$$
on $\mathbb{R}^{d+m}$. We apply the Skorohod representation theorem, as it ensures that convergence in distribution can be realized almost surely by mapping a single uniform random variable through an appropriate sequence of measurable functions. Specifically, there exist a random variable $U_1 \sim \text{Uniform}(0,1)$ and measurable functions $A,A_n: [0,1] \to \mathbb{R}^{d+m}$ such that for all $n$: 
(i) $A_n(U_1) \overset{d}{=} \left( n^a(\hat{\theta}_n (D_n) - \theta_0), g_n(D_n)\right)^\top$ for each $n$, 
(ii) $A(U_1) \overset{d}{=} (X, c)^\top$, 
(iii) $A_n(U_1) \overset{a.s.}{\to} A(U_1)$. 
In other words, let $E_1 = \{u_1 \in [0,1]: A_n(u_1) \to A(u_1)\}$. We have $P(U_1 \in E_1)=1$.

Second, we construct the synthetic data through representation. According to Theorem 10.2.2 of \citet{dudley2002real}, in a probability space $(\mathcal{S^\infty, \mathcal{F}^\infty, \mathcal{Q}}^\infty)$, if, for any given $n$, $\mathcal{S}^n$ is a polish space with a Borel algebra $\sigma(\mathcal{S}^n)$, $\text{proj}_n: \mathcal{S}^\infty \to \mathcal{S}^n$ is a measurable function and $\sigma(\hat{\theta}_n) \subseteq \mathcal{F}^\infty$ is a sub $\sigma$ algebra, then there exists a unique (up to a measure 0 set in $\mathcal{S}^\infty$) conditional distribution $Q^n (\cdot \mid \hat{\theta}_n)$ on $\mathcal{S}^n$.

Fix for any $u_1 \in E_1$. Let the synthetic $\tilde{\theta}_n (u_1, \theta_0) \defn \theta_0 + A_{n, 1:d}(u_1)/n^a$, which has the same law as $\hat{\theta}_n(D_n)$ over the randomness of $U_1$. Since the conditional distributions exist for each $n$, we can define random variables $\tilde{D}_n^{u_1} \in \mathcal{S}^n$ with the following properties: 
(i) $\hat{\theta}_n(\tilde{D}_n^{u_1})=\tilde{\theta}_n (u_1, \theta_0)$ and $g_n(\tilde{D}_n^{u_1}) = A_{n, (d+1): (d+m)}(u_1)$,
(ii) Since $u_1 \in E_1$ and $g_n(\tilde{D}_n^{u_1}) {=} A_{n,(d+1,d+m)} (u_1) \to A(u_1)  = c$, $(\tilde{D}_n^{u_1})_{n=1}^\infty \in G$,
(iii) $\tilde{D}_n^{U_1} \overset{d}{=} D_n$.


Now, we constructing the (independent) mechanism error via $U_2$.
We can again apply the Skorohod representation theorem to simulate the conditional convergence. Fix for any $u_1 \in E_1$ and $(\tilde{\mathbf{d}}_n^{u_1})_{n=1}^\infty \in G$. By (B2) and Skorohod,  there exists $U_2 \sim \text{Uniform}(0,1)$, constructed independently of $U_1$, and measurable functions $B,B_{n,A_n(u_1),(\tilde{\mathbf{d}}_n^{u_1})_{n=1}^\infty}: [0,1] \to \mathbb{R}^d$ such that  for all $n$:
(i) $B_{n,A_n(u_1),(\tilde{\mathbf{d}}_n^{u_1})_{n=1}^\infty} (U_2) \overset{d}{=} n^b (\theta_n^{\mathrm{mech}}(D_n) - \hat{\theta}_n(D_n))$,
(ii) $B(U_2) \overset{d}{=} Y$, 
(iii) $B_{n,A_n(u_1),(\tilde{\mathbf{d}}_n^{u_1})_{n=1}^\infty} (U_2) \overset{a.s.}{\to} B(U_2)$. 
Let the synthetic $\tilde{\theta}_n^\text{mech} (u_1,u_2,\tilde{\mathbf{d}}_n^{u_1}, \theta_0) \defn \tilde{\theta}_n (u_1, \theta_0) + B_{n,A_n(u_1),(\tilde{d}_n^{u_1})_{n=1}^\infty} (u_2) / n^b$, which has the same law as $\theta_n^{\text{mech}}(D_n)$ over the randomness of $U_1,U_2$. Let $E_2 = \{(u_1,u_2) \in [0,1]^2: A_n(u_1) \to A(u_1), B_{n,A_n(u_1),(\tilde{D}_n^{u_1})_{n=1}^\infty} (u_2) \to B(u_2)\}$. We have $P((U_1,U_2) \in E_2)=1$.

We have derived respectively that $\tilde{D}_n^{U_1} \overset{d}{=} D_n$, and $\tilde{\theta}_n(U_1,\theta_0) \overset{d}{=} \hat{\theta}_n(D_n)$ and, for all $(\tilde{\mathbf{d}}_n^{u_1})_{n=1}^\infty \in G$, $\tilde{\theta}_n^\text{mech} (u_1,U_2,\tilde{\mathbf{d}}_n^{u_1}, \theta_0) \overset{d}{=} \theta_n^{\text{mech}}(\tilde{\mathbf{d}}_n^{u_1})$. This implies that 
\begin{align} \label{eq: joint dist over data and mechanism randomness}
\begin{pmatrix}
    \tilde{D}_n^{U_1} \\ \tilde{\theta}_n(U_1,\theta_0) \\\tilde{\theta}_n^\text{mech}(U_1,U_2,\tilde{D}_n^{U_1}, \theta_0)
\end{pmatrix}    
\overset{d}{=}
\begin{pmatrix}
    D_n \\ \hat{\theta}_n(D_n) \\\theta_n^\text{mech}(D_n)
\end{pmatrix}.
\end{align}

Now, from the construction above, we have the following:
$$
\begin{pmatrix}
n^a (\hat{\theta}_n(D_n) - \theta_0)\\
n^b \bigl(\theta_n^{\mathrm{mech}}(D_n) - \hat{\theta}_n(D_n)\bigr)
\end{pmatrix}
\overset{d}{=}
\begin{pmatrix}
A_{n,1:d} (U_1) \\ B_{n,A_n(U_1),(\tilde{D}_n^{U_1})_{n=1}^\infty} (U_2)
\end{pmatrix}
\overset{a.s.}{\to}
\begin{pmatrix}
A_{1:d}(U_1) \\ B(U_2)
\end{pmatrix}
\overset{d}{=}
\begin{pmatrix}
X \\ Y
\end{pmatrix},
$$
where the first equality in distribution holds from \eqref{eq: joint dist over data and mechanism randomness}, the joint almost sure convergence holds from stacking individual almost surely convergent sequence, and the last equality in distribution holds as $U_1$ and $U_2$ are independent.
\end{proof}

\emph{Normalized-Statistic Form.} The proof above uses the normalized errors only through their joint law with the dataset, so it applies without change to more general statistics. Let $T_n:\mathcal S^n\to\mathbb R^d$ be measurable with $T_n(D_n)\dlim X$, and let $W_n:\mathbb R^d\times\mathcal S^n\to\mathbb R^d$ be measurable such that, for every fixed sequence $(\mathbf s_n)_{n\ge1}\in G$, $W_n(\theta_n^{\mathrm{mech}}(\mathbf s_n),\mathbf s_n)\dlim Y$ over the mechanism randomness, where the law of $Y$ is the same for every sequence in $G$. Replacing $n^a(\hat\theta_n-\theta_0)$ by $T_n$ and $n^b(\theta_n^{\mathrm{mech}}-\hat\theta_n)$ by $W_n$ throughout the proof gives $(T_n(D_n),W_n(\theta_n^{\mathrm{mech}}(D_n),D_n))\dlim(X,Y)$, with $X$ and $Y$ independent. This form allows deterministic normalizing sequences that are not powers of $n$ and centers other than $\hat\theta_n$. In particular, Section~\ref{sec: joint exp mech} uses $T_n=\sqrt n(\hat\theta_n-\theta_0)$ together with $W_n=\epsilon_n n(\theta_n^{DP}-\hat\theta_n)$, $W_n=\epsilon_n\sqrt{n\mu_n}(\theta_n^{DP}-\theta_n^\#)$, or $W_n=\epsilon_n\sqrt{n\mu_n}\{(I_d+(\mu_n/n)H^{-1})\theta_n^{DP}-\hat\theta_n\}$.

\restatePropAsRegularity*
\begin{proof}
We verify (A2)--(A4) on a common probability-one event; (A1) is already assumed. By (A2$'$), for $\Pi$-almost every record $s$, the loss $f_\theta(s)$ is twice continuously differentiable on the common neighborhood $\mathcal U$ and satisfies
\[
\|\nabla^2f_\theta(s)-\nabla^2f_{\theta'}(s)\|
\le M(s)\|\theta-\theta'\|,
\qquad \theta,\theta'\in\mathcal U.
\]
Each sampled record satisfies these conditions with probability one. Taking the countable intersection over $i$ shows that they hold simultaneously for every $s_i$ almost surely. Moreover, $\mathbb E[M(S)]<\infty$ by (A2$'$), so the SLLN gives
\[
\overline M_n=\frac1n\sum_{i=1}^nM(s_i)\to\mathbb E[M(S)]<\infty
\quad\text{almost surely}.
\]
Since every convergent real sequence is bounded, this implies $\sup_n\overline M_n<\infty$ almost surely. Finally, the $L$-Lipschitz condition on $\theta\mapsto f_\theta(s)-f_\theta(s')$ for every pair $s,s'\in\mathcal S$ is explicitly assumed in (A2$'$). These observations verify all requirements of (A2) on a common probability-one event.

To verify (A3), we need an almost-sure convergence rate for the empirical Hessian at $\theta_0$. We also establish the corresponding gradient rate, which will be used to verify (A4).

We first justify that the gradient at $\theta_0$ has mean zero. Choose a closed ball $\overline B(\theta_0,r)\subset\mathcal U$. The fundamental theorem of calculus gives $\nabla f_\theta(S)-\nabla f_{\theta_0}(S)
=\int_0^1\nabla^2f_{\theta_0+t(\theta-\theta_0)}(S)(\theta-\theta_0)\,dt$. For every $\theta$ in this ball, the Hessian Lipschitz bound implies
\[
\begin{aligned}
\|\nabla^2f_{\theta_0+t(\theta-\theta_0)}(S)\|
&\le \|\nabla^2f_{\theta_0}(S)\|
+\|\nabla^2f_{\theta_0+t(\theta-\theta_0)}(S)-\nabla^2f_{\theta_0}(S)\|\\
&\le \|\nabla^2f_{\theta_0}(S)\|
+tM(S)\|\theta-\theta_0\|.
\end{aligned}
\]
Consequently,
\[
\begin{aligned}
\|\nabla f_\theta(S)\|
&\le \|\nabla f_{\theta_0}(S)\|
+\|\theta-\theta_0\|\int_0^1
\|\nabla^2f_{\theta_0+t(\theta-\theta_0)}(S)\|\,dt\\
&\le \|\nabla f_{\theta_0}(S)\|
+\|\theta-\theta_0\|\|\nabla^2f_{\theta_0}(S)\|
+\frac12M(S)\|\theta-\theta_0\|^2\\
&\le \|\nabla f_{\theta_0}(S)\|
+r\|\nabla^2f_{\theta_0}(S)\|+\frac{r^2}{2}M(S).
\end{aligned}
\]
This bound holds uniformly over the ball. Its right-hand side has finite expectation, since (A2$'$)--(A3$'$) and Cauchy--Schwarz give
\[
\begin{aligned}
\mathbb E\|\nabla f_{\theta_0}(S)\|
&\le \bigl(\mathbb E\|\nabla f_{\theta_0}(S)\|^2\bigr)^{1/2}
=\sqrt{\operatorname{tr}(\mathcal I)}<\infty,\\
\mathbb E\|\nabla^2f_{\theta_0}(S)\|
&\le \bigl(\mathbb E\|\nabla^2f_{\theta_0}(S)\|^2\bigr)^{1/2}<\infty,
\qquad \mathbb E[M(S)]<\infty.
\end{aligned}
\]
The preceding integrable bound justifies interchanging differentiation and expectation. Since $\theta_0$ is an interior minimizer of $F$, its gradient vanishes, and hence
\[
\mathbb E[\nabla f_{\theta_0}(S)]
=\left.\nabla_\theta\mathbb E[f_\theta(S)]\right|_{\theta=\theta_0}
=\nabla F(\theta_0)
=0.
\]

Both the empirical Hessian and gradient are averages of i.i.d. quantities with finite second moments:
\[
\nabla^2F_n(\theta_0)=\frac1n\sum_{i=1}^n\nabla^2f_{\theta_0}(s_i),
\qquad
\nabla F_n(\theta_0)=\frac1n\sum_{i=1}^n\nabla f_{\theta_0}(s_i).
\]
We establish a common rate for scalar averages, then apply it to each Hessian entry and gradient coordinate. Let $V_i$ be generic i.i.d. scalar random variables with finite second moment. For every $\zeta>0$, we claim that
\begin{equation}\label{eq: strong law with slack}
n^{1/2-\zeta}\left|\frac1n\sum_{i=1}^nV_i-\mathbb EV_1\right|\to0
\quad\text{almost surely}.
\end{equation}
Write $R_\ell=\sum_{i=1}^{\ell}(V_i-\mathbb EV_1)$. Kolmogorov's maximal inequality gives, for every $a>0$,
\[
\Pr\left(\max_{1\le\ell\le2^k}|R_\ell|>a\,2^{k(1/2+\zeta)}\right)
\le \frac{2^k\operatorname{Var}(V_1)}{a^2\,2^{k(1+2\zeta)}}
=\frac{\operatorname{Var}(V_1)}{a^2}2^{-2k\zeta}.
\]
Since the upper bounds are summable, Borel--Cantelli with $a=1/q$ gives 
$$
\Pr\left(
\frac{\max_{\ell\le2^k}|R_\ell|}{2^{k(1/2+\zeta)}}>\frac1q\text{ infinitely often}\right)=0.
$$ 
Thus, almost surely, $\frac{\max_{\ell\le2^k}|R_\ell|}{2^{k(1/2+\zeta)}}\le\frac1q$ for all sufficiently large $k$. Taking a countable intersection over $q$ and letting $q\to\infty$ shows that this ratio converges to zero. For $2^{k-1}<n\le2^k$,
\[
\begin{aligned}
n^{1/2-\zeta}\left|\frac1n\sum_{i=1}^nV_i-\mathbb EV_1\right|
=\frac{|R_n|}{n^{1/2+\zeta}}\le\frac{\max_{\ell\le2^k}|R_\ell|}{2^{(k-1)(1/2+\zeta)}}
=2^{1/2+\zeta}\frac{\max_{\ell\le2^k}|R_\ell|}{2^{k(1/2+\zeta)}}\to0,
\end{aligned}
\]
proving \eqref{eq: strong law with slack}.

For each Hessian entry, take $V_i=[\nabla^2f_{\theta_0}(s_i)]_{jk}$, whose expectation is $H_{jk}$ and whose second moment is finite by (A3$'$). For each gradient coordinate, take $V_i=[\nabla f_{\theta_0}(s_i)]_j$, whose expectation is zero as established above and whose second moment is finite by (A2$'$). Applying \eqref{eq: strong law with slack} to these finitely many entries gives
\begin{equation}\label{eq: strong rates at theta0}
n^{1/2-\zeta}\|\nabla^2F_n(\theta_0)-H\|\to0,
\qquad
n^{1/2-\zeta}\|\nabla F_n(\theta_0)\|\to0
\end{equation}
almost surely, since coordinatewise convergence implies norm convergence in fixed dimension. The first limit verifies (A3); the second will give the rate in (A4).

To verify (A4), local Hessian smoothness gives a fixed ball $K=\overline B(\theta_0,r_0)\subset\mathcal U$ and $m>0$ such that $\nabla^2F_n(\theta)\succeq mI_d$ on $K$ eventually. For any fixed $0<\delta\le r_0$ and $\|\theta-\theta_0\|=\delta$,
\[
\begin{aligned}
F_n(\theta)-F_n(\theta_0)
\ge \nabla F_n(\theta_0)^\top(\theta-\theta_0) +\frac m2\|\theta-\theta_0\|^2
\ge -\|\nabla F_n(\theta_0)\|\delta+\frac m2\delta^2>0
\end{aligned}
\]
eventually, because $\nabla F_n(\theta_0)\to0$. Continuity and compactness ensure that the minimum over $\overline B(\theta_0,\delta)$ is attained, and the positive boundary gap places it in the interior. Its gradient is zero, so global convexity makes it a global minimizer. Note that no global minimizer, say $\theta^*$ can lie outside the ball; this is because convexity would then produce a boundary point, which is between $\theta^*$ and $\theta_0$, with objective value less than or equal to $F_n(\theta_0)$, causing contradiction. Local positive curvature ensures uniqueness.

The argument above shows that, for every fixed $0<\delta\le r_0$, the minimizer $\hat\theta_n$ of $F_n$ eventually satisfies $\|\hat\theta_n-\theta_0\|<\delta$. We apply the same argument separately to the minimizer $\theta_n^\#$ of $F_n^\#$. Since $\mu_n=o(n)$,
\[
\nabla F_n^\#(\theta_0)
=\nabla F_n(\theta_0)+\frac{\mu_n}{n}\theta_0\to0,
\qquad
\nabla^2F_n^\#(\theta)\succeq mI_d
\quad\text{on }K.
\]
Thus, for each fixed $0<\delta\le r_0$, the boundary gap and convexity argument applied to $F_n^\#$ gives $\|\theta_n^\#-\theta_0\|<\delta$ eventually. Since $\delta$ can be arbitrarily small, these two conclusions prove $\hat\theta_n\to\theta_0$ and $\theta_n^\#\to\theta_0$ on the probability-one event.

For the minimizer rate, the first-order condition and the fundamental theorem of calculus give
\[
0=\nabla F_n(\hat\theta_n)
=\nabla F_n(\theta_0)
+\left[\int_0^1\nabla^2F_n\bigl(\theta_0+t(\hat\theta_n-\theta_0)\bigr)\,dt\right](\hat\theta_n-\theta_0).
\]
By consistency, the segment eventually lies in $K$, so the integrated Hessian is bounded below by $mI_d$ and its inverse has norm at most $1/m$. Consequently,
\[
n^{1/2-\zeta}\|\hat\theta_n-\theta_0\|
\le \frac1m n^{1/2-\zeta}\|\nabla F_n(\theta_0)\|\to0
\quad\text{almost surely}.
\]
Together with regularized consistency, this verifies (A4).

Measurable minimizers can be selected on a fixed compact ball about $\theta_0$ contained in $\mathcal U$; these eventually coincide with the unique global minimizers. Intersecting the probability-one events above establishes (A1)--(A4) simultaneously, proving $\Pi^\infty(G)=1$. On this event, the preceding limits for $\overline M_n$, $\nabla^2F_n(\theta_0)$, $\hat\theta_n$, and $\theta_n^\#$ also give $g_n(D_n)\to c$.
\end{proof}

\restateThmExpMechJointGeneral*
\begin{proof}
Let $G$ be defined by \eqref{eq: almost sure data sequences}. Proposition~\ref{prop: as regularity} gives $\Pi^\infty(G)=1$, and (A4$'$) gives the data-side limit $X_n=\sqrt n(\hat\theta_n-\theta_0)\dlim N_d(0,\Sigma_1)$. The conditional results of Section~\ref{sec: dp error asymp} give a fixed Gaussian mechanism limit for every sequence in $G$. Applying the stacking argument, in its normalized-statistic form explained after its proof, yields
\begin{equation}\label{eq: centered joint limit}
\begin{pmatrix}X_n\\Z_n\end{pmatrix}\dlim N_{2d}\left(0,\begin{pmatrix}\Sigma_1&0\\0&C\end{pmatrix}\right),
\end{equation}
where $Z_n=\epsilon_n n(\theta_n^{DP}-\hat\theta_n)$ and $C=(L^2/\mu)H^{-1}$ in the strongly convex case, while $Z_n=\epsilon_n\sqrt{n\mu_n}(\theta_n^{DP}-\theta_n^\#)$ and $C=L^2H^{-1}$ in the convex case. This establishes the strongly convex assertion and the centered joint limit needed for all the convex cases.

\emph{Cases (i) and (ii).}
Lemma~\ref{lem: regularized convex utility} holds on $G$, so $B_n=(n/\mu_n)(\theta_n^\#-\hat\theta_n)\to-H^{-1}\theta_0$ almost surely. For (i), apply joint Slutsky to $(X_n,Z_n+\kappa_n B_n)$ when $\kappa_n\to\kappa<\infty$. For (ii), apply it to $(X_n,Z_n/\kappa_n+B_n)$ when $\kappa_n\to\infty$. The second component then converges to a constant; the first retains its sampling limit.

\emph{Case (iii): Oracle Bias Correction.}
Write $\lambda_n=\mu_n/n$ and $A_n=I_d+(\mu_n/n)H^{-1}$. Under \eqref{eq: conditional correction tuning}, Corollary~\ref{cor: conditional bias correction} gives a centered conditional Gaussian limit for $\epsilon_n\sqrt{n\mu_n}(A_n\theta_n^{DP}-\hat\theta_n)$ on every sequence in $G$. Stacking this law with the data limit proves (iii) directly.

\end{proof}

\restateCorHessianStability*
\begin{proof}
Let $\Delta_n=\theta_n^\#-\hat\theta_n$ and $H_n=\nabla^2F_n(\hat\theta_n)$. Lemma~\ref{lem: regularized convex utility} on the probability-one set $G$ gives $(n/\mu_n)\Delta_n\to-H^{-1}\theta_0$ almost surely, hence $\Delta_n=O_p(\mu_n/n)$, and $H_n\to H\succ0$ in probability. All inverse identities below are used on the event that $H_n$ is invertible, whose probability tends to one; define the corrected statistic arbitrarily on the complement.

The first-order conditions and integral Taylor expansion give
\begin{equation}\label{eq: integral form of Taylor expansion}
\left(H_n+\frac{\mu_n}{n}I_d\right)\Delta_n=-\frac{\mu_n}{n}\hat\theta_n-r_n,\qquad r_n=\int_0^1[\nabla^2F_n(\hat\theta_n+t\Delta_n)-H_n]\Delta_n\,dt.
\end{equation}
By definition of $\rho_n$, $\|r_n\|\le\rho_n\|\Delta_n\|=O_p(\rho_n\mu_n/n)$. Writing $\theta_n^{DP}=\theta_n^\#+(\theta_n^{DP}-\theta_n^\#)$ gives
\begin{align}
\widetilde\theta_n-\hat\theta_n&=\left(I_d+\frac{\mu_n}{n}H_n^{-1}\right)(\theta_n^{DP}-\theta_n^\#)+\Delta_n+\frac{\mu_n}{n}H_n^{-1}\theta_n^\#\nonumber\\
&=\left(I_d+\frac{\mu_n}{n}H_n^{-1}\right)(\theta_n^{DP}-\theta_n^\#)-H_n^{-1}r_n.\label{gap between tilde and hat based on bias correction}
\end{align}
The last equality follows by multiplying \eqref{eq: integral form of Taylor expansion} by $H_n^{-1}$. The stability condition gives
\[
\epsilon_n\sqrt{n\mu_n}\|H_n^{-1}r_n\|=O_p\left(\frac{\epsilon_n\mu_n^{3/2}\rho_n}{\sqrt n}\right)=o_p(1).
\]
Since $(\mu_n/n)H_n^{-1}\to0$ in probability and the centered mechanism component in \eqref{eq: centered joint limit} is tight, the normalized corrected error differs from that component by $o_p(1)$. Joint Slutsky proves the corollary.

Finally, (A2$'$) gives $\rho_n\le\overline M_n\|\Delta_n\|=O_p(\mu_n/n)$, so $\epsilon_n\mu_n^{5/2}=o(n^{3/2})$ is sufficient. A Hessian constant along the segment gives $r_n=0$ exactly. If the segment lies in a quadratic neighborhood with probability tending to one, the same conclusion follows because the exceptional event has vanishing probability.
\end{proof}

\restateThmConfidenceRegionGeneral*
\begin{proof}
Write $X_n=\sqrt n(\hat\theta_n-\theta_0)$ and $\Sigma_1=H^{-1}\mathcal I H^{-1}\succ0$. We follow the strongly convex case and the two convex centering choices separately, then give the common studentization argument.

\emph{Strongly Convex Loss Case.}
Put $Z_n=\epsilon_n n(\theta_n^{DP}-\hat\theta_n)$ and $C=(L^2/\mu)H^{-1}$. The joint theorem gives independent centered Gaussian limits for $(X_n,Z_n)$ with covariances $\Sigma_1,C$. The exact decomposition is
\[
\sqrt n(\theta_n^{DP}-\theta_0)=X_n+\frac{Z_n}{\epsilon_n\sqrt n}.
\]
Thus the covariance combination is $V_n=\Sigma_1+L^2H^{-1}/(\mu\epsilon_n^2n)$, estimated by the stated $\hat V_n$. Its second term may grow when $\epsilon_n$ decreases. The normalized argument below allows this growth; boundedness of $V_n$ is not required.

\emph{Convex Loss Case.}
Let $Z_n=\epsilon_n\sqrt{n\mu_n}(\theta_n^{DP}-\theta_n^\#)$ and $C=L^2H^{-1}$. Equation~\eqref{eq: centered joint limit} gives independent centered Gaussian limits for $(X_n,Z_n)$ with covariances $\Sigma_1,C$.

We first derive stochastic bias bounds under the general RegExpMech assumptions, without imposing \eqref{eq: conditional correction tuning}. By (A3$'$), Chebyshev's inequality gives $\|\nabla^2F_n(\theta_0)-H\|=O_p(n^{-1/2})$; (A4$'$) gives $\hat\theta_n-\theta_0=O_p(n^{-1/2})$, and the SLLN gives $\overline M_n=O_p(1)$. Let $\Delta_n=\theta_n^\#-\hat\theta_n$ and $J_n=\int_0^1\nabla^2F_n(\hat\theta_n+t\Delta_n)\,dt$. Consistency implies $J_n\to H$ in probability, so \eqref{eq: deterministic bias identity} gives $\Delta_n=O_p(\mu_n/n)$. Consequently,
\[
\|J_n-H\|
\le\|\nabla^2F_n(\theta_0)-H\|
+\overline M_n(\|\hat\theta_n-\theta_0\|+\|\Delta_n\|)
=O_p\left(n^{-1/2}+\frac{\mu_n}{n}\right).
\]
The matrix inverse identity applied to
$\Delta_n=-(\mu_n/n)(J_n+(\mu_n/n)I_d)^{-1}\hat\theta_n$ yields
\begin{equation}\label{eq: stochastic bias expansion}
\begin{aligned}
\theta_n^\#-\hat\theta_n
&=-\frac{\mu_n}{n}H^{-1}\theta_0
+O_p\left(\frac{\mu_n}{n^{3/2}}+\frac{\mu_n^2}{n^2}\right),\\
\theta_n^\#-\theta_0
&=O_p\left(n^{-1/2}+\frac{\mu_n}{n}\right).
\end{aligned}
\end{equation}
For $A_n=I_d+(\mu_n/n)H^{-1}$, these bounds give
\begin{equation}\label{eq: oracle stochastic correction}
\begin{aligned}
A_n\theta_n^{DP}-\hat\theta_n
&=A_n(\theta_n^{DP}-\theta_n^\#)
+\Delta_n+\frac{\mu_n}{n}H^{-1}\theta_n^\#\\
&=A_n(\theta_n^{DP}-\theta_n^\#)
+O_p\left(\frac{\mu_n}{n^{3/2}}+\frac{\mu_n^2}{n^2}\right),
\end{aligned}
\end{equation}
since the leading terms $-(\mu_n/n)H^{-1}\theta_0$ and $(\mu_n/n)H^{-1}\theta_0$ cancel. We now compare these remainders with the combined error scale $s_n$.

\emph{Case 1: Uncorrected Center.}
The exact decomposition is
\[
\sqrt n(\theta_n^{DP}-\theta_0)=X_n+\frac{Z_n}{\epsilon_n\sqrt{\mu_n}}+\sqrt n(\theta_n^\#-\hat\theta_n).
\]
By \eqref{eq: stochastic bias expansion}, $\theta_n^\#-\hat\theta_n=O_p(\mu_n/n)$. Hence $(\theta_n^\#-\hat\theta_n)/s_n=o_p(1)$ under the uncorrected centering condition. The covariance combination is $V_n=\Sigma_1+L^2H^{-1}/(\epsilon_n^2\mu_n)$.

\emph{Case 2: Bias-Corrected Center.}
First use the population correction $A_n=I_d+(\mu_n/n)H^{-1}$. Equation~\eqref{eq: oracle stochastic correction} gives
\[
\sqrt n(A_n\theta_n^{DP}-\theta_0)=X_n+\frac{A_nZ_n}{\epsilon_n\sqrt{\mu_n}}+O_p\left(\frac{\mu_n}{n}+\frac{\mu_n^2}{n^{3/2}}\right).
\]
After division by $\sqrt n\,s_n$, the remainder tends to zero: its first term is bounded by $O_p(\mu_n/n)$ because $s_n\ge n^{-1/2}$, and its second vanishes by $\mu_n^2/(n^2s_n)\to0$. Replacing $A_nZ_n$ by $Z_n$ also adds $o_p(1)$ on this scale, since $1/(\epsilon_n\sqrt{n\mu_n}s_n)\le1$ and $A_n\to I_d$.

It remains to replace $A_n$ by $\hat A_n$. Concentration and consistency give $\theta_n^{DP}=\theta_n^\#+Z_n/(\epsilon_n\sqrt{n\mu_n})\to\theta_0$ in probability. Therefore
\[
\frac{\|(\hat A_n-A_n)\theta_n^{DP}\|}{s_n}\le\frac{\mu_n}{ns_n}\|\hat\Sigma_{2,n}-H^{-1}\|\,\|\theta_n^{DP}\|=o_p(1)
\]
by \eqref{eq: private correction rate}. This verifies the centering condition for the estimated correction, without assuming it is independent of the mechanism draw.

\emph{Covariance Standardization and Coverage.}
To handle all privacy sequences, including ones for which the mechanism variance dominates or its relative weight oscillates, put $t_n=1/(\epsilon_n\sqrt n)$ in the strongly convex case and $t_n=1/(\epsilon_n\sqrt{\mu_n})$ in the convex case, and let $q_n=\sqrt{1+t_n^2}$. The preceding calculations give
\begin{equation}\label{eq: normalized total error}
\frac{\sqrt n(\tilde\theta_n-\theta_0)}{q_n}=\frac{X_n+t_nZ_n}{q_n}+o_p(1).
\end{equation}
Also $\hat V_n=\hat\Sigma_{1,n}+t_n^2\hat C_n$, where $\hat C_n=(L^2/\mu)\hat\Sigma_{2,n}$ or $L^2\hat\Sigma_{2,n}$, respectively, and $\hat C_n\to C\succ0$ in probability.

Every subsequence has a further subsequence on which $u_n=1/q_n\to u$ and $v_n=t_n/q_n\to v$, with $u,v\ge0$ and $u^2+v^2=1$. On that further subsequence, joint convergence and \eqref{eq: normalized total error} give a centered Gaussian limit with covariance $u^2\Sigma_1+v^2C\succ0$, while
\[
\frac{\hat V_n}{q_n^2}=u_n^2\hat\Sigma_{1,n}+v_n^2\hat C_n\xrightarrow{P}u^2\Sigma_1+v^2C.
\]
The continuous mapping theorem for the inverse square root and Slutsky's theorem give $\sqrt n\,\hat V_n^{-1/2}(\tilde\theta_n-\theta_0)\dlim N_d(0,I_d)$ on every such further subsequence, and hence on the full sequence. This argument also proves that $\hat V_n$ is positive definite with probability tending to one; define the statistic arbitrarily otherwise. Finally,
\[
n(\tilde\theta_n-\theta_0)^\top\hat V_n^{-1}(\tilde\theta_n-\theta_0)\dlim\chi_d^2,
\]
which proves the confidence-region coverage.
\end{proof}

\emph{Recovery of the non-private limit.} For RegExpMech, $\epsilon_n\sqrt{\mu_n}\to\infty$ makes mechanism noise negligible relative to sampling error. Under this condition, \eqref{eq: stochastic bias expansion} gives
\[
\sqrt n(\theta_n^{DP}-\theta_0)\dlim N_d(0,\Sigma_1)
\qquad\text{if }\mu_n=o(n^{1/2}).
\]
For the oracle correction $A_n=I_d+(\mu_n/n)H^{-1}$,
\eqref{eq: oracle stochastic correction} gives
\[
\sqrt n(A_n\theta_n^{DP}-\theta_0)\dlim N_d(0,\Sigma_1)
\qquad\text{if }\mu_n=o(n^{3/4}).
\]
Replacing $H^{-1}$ by a private estimate $\hat\Sigma_{2,n}$ additionally requires
$\mu_n\|\hat\Sigma_{2,n}-H^{-1}\|=o_p(\sqrt n)$.
These sufficient conditions recover the non-private limiting distribution. The confidence-region theorem also provides valid inference when mechanism noise is comparable to or larger than sampling error, provided its centering and covariance-estimation conditions hold.

\section{Proofs of Main-Text Specializations}\label{app: specialization proofs}

The main text states the strongly convex results, the balanced regime, and the fixed-budget cases. This section derives those statements from the general results proved above.

\begin{proof}[Proof of \eqref{eq: balanced privacy normalization}]
Take $\mu_n=(n/\epsilon_n^2)^{1/3}$. Then $\kappa_n=1$, $\epsilon_n\sqrt{n\mu_n}=(\epsilon_n n)^{2/3}\to\infty$, and $\mu_n/n=(\epsilon_n n)^{-2/3}\to0$. Part (ii) of Theorem~\ref{thm: regularized convex utility} therefore gives \eqref{eq: balanced privacy normalization}.
\end{proof}

\restateCorBalancedTuning*
\begin{proof}
Let $\epsilon_n=\epsilon>0$. In case (i), $\epsilon\sqrt{n\mu_n}=\epsilon n^r\to\infty$, $\mu_n=o(n)$, and $\kappa_n=\epsilon n^{3r-2}\to0$ because $r<2/3$. Part (i) of Theorem~\ref{thm: regularized convex utility}, divided by $\epsilon$, gives the stated limit. In case (ii), $\epsilon\sqrt{n\mu_n}=\epsilon n^{2/3}\to\infty$, $\mu_n=o(n)$, and $\kappa_n=\epsilon$. Part (ii) of the theorem, divided by $\epsilon$, gives the stated limit. In case (iii), $\epsilon\sqrt{n\mu_n}=\epsilon n^{1-r/2}\to\infty$, $\mu_n=o(n)$ because $r>0$, and $\kappa_n=\epsilon n^{1-3r/2}\to\infty$ because $r<2/3$. Since $n/\mu_n=n^r$, part (iii) of the theorem gives the stated limit.
\end{proof}

\begin{proof}[Proof of the Balanced and Fixed-Budget Claims for Corollary~\ref{cor: conditional bias correction}]
If $\kappa_n\to\kappa\in(0,\infty)$, then $\epsilon_n\mu_n^{3/2}=O(n^{1/2})=o(n^{1-\zeta})$ because $\zeta<1/10$, and $\epsilon_n\mu_n^{5/2}=O(n^{1/2}\mu_n)=o(n^{3/2})$ because $\mu_n=o(n)$. Thus \eqref{eq: conditional correction tuning} holds in the balanced regime.

Now let $\epsilon_n=\epsilon>0$ and $\mu_n=n^\gamma$ with $0<\gamma<3/5$. Then $\epsilon\sqrt{n\mu_n}=\epsilon n^{(1+\gamma)/2}\to\infty$ and $\mu_n=o(n)$. Also, $\epsilon\mu_n^{3/2}=\epsilon n^{3\gamma/2}=o(n^{9/10})=o(n^{1-\zeta})$ because $\gamma<3/5$ and $\zeta<1/10$, and $\epsilon\mu_n^{5/2}=\epsilon n^{5\gamma/2}=o(n^{3/2})$ because $\gamma<3/5$. Corollary~\ref{cor: conditional bias correction}, divided by $\epsilon$, gives the stated fixed-budget limit.
\end{proof}

\restateThmExpMechJoint*
\begin{proof}
Part 1 is the strongly convex statement of Theorem~\ref{thm: exp mech joint general}. For Part 2, the preceding proof shows that $\kappa_n\to\kappa\in(0,\infty)$ implies \eqref{eq: conditional correction tuning}. Part (iii) of Theorem~\ref{thm: exp mech joint general} therefore gives the stated limit, with $\widetilde\theta_n=A_n\theta_n^{DP}$.
\end{proof}

\emph{Squared-Error Example for Corollary~\ref{cor: hessian stability}.} Consider squared-error linear regression,
\[
F_n(\theta)=\frac1{2n}\sum_{i=1}^n(y_i-x_i^\top\theta)^2.
\]
Its Hessian is $\nabla^2F_n(\theta)=n^{-1}\sum_{i=1}^n x_ix_i^\top$, independent of $\theta$, so $\rho_n=0$ and the Hessian stability condition holds automatically. Provided the remaining assumptions hold, the corollary therefore applies whenever $\mu_n=o(n)$ and $\epsilon_n\sqrt{n\mu_n}\to\infty$. In particular, for a fixed positive privacy budget, $\mu_n=n^{4/5}$ is allowed, although it lies outside the generic first-order correction range $\mu_n=o(n^{3/5})$.

\restateThmConfidenceRegion*
\begin{proof}
Part 1 is the strongly convex statement of Theorem~\ref{thm: confidence region general}. For Part 2, the definition of $s_n$ in \eqref{eq: sampling comparison scales} gives $s_n\ge1/(\epsilon_n\sqrt{n\mu_n})$. Consequently,
\[
\frac{\mu_n}{ns_n}\le\kappa_n,
\qquad
\frac{\mu_n^2}{n^2s_n}\le\frac{\mu_n}{n}\kappa_n\to0.
\]
Since $\kappa_n$ is bounded and $\hat\Sigma_{2,n}$ is consistent,
\[
\frac{\mu_n}{s_n}\|\hat\Sigma_{2,n}-H^{-1}\|\le n\kappa_n\|\hat\Sigma_{2,n}-H^{-1}\|=o_p(n).
\]
Thus the bias-corrected case of \eqref{eq: private correction rate} holds, and Theorem~\ref{thm: confidence region general} gives the stated result.
\end{proof}

\begin{proof}[Proof of Remark~\ref{rem: fixed privacy inference}]
Let $\epsilon_n=\epsilon>0$. In the strongly convex case, $\epsilon n\to\infty$, so Part 1 of Theorem~\ref{thm: confidence region} applies. In the regularized case, $\mu_n=n^{1/3}$ gives $\mu_n=o(n)$, $\epsilon\sqrt{n\mu_n}=\epsilon n^{2/3}\to\infty$, and $\kappa_n=\epsilon$, so Part 2 applies. Substituting $\epsilon_n=\epsilon$ and $\mu_n=n^{1/3}$ gives the displayed $\tilde\theta_n$ and $\hat V_n$.
\end{proof}

\bibliography{bibliography}

\end{document}